\documentclass[a4paper,fleqn]{cas-dc}

\usepackage[T1]{fontenc}
\usepackage{lmodern}
\usepackage[numbers,sort&compress]{natbib}
\usepackage{array}
\usepackage{adjustbox}
\usepackage{longtable}
\usepackage{caption}
\usepackage{xcolor}


\usepackage{multirow}
\usepackage{makecell}
\usepackage{threeparttable}
\newcommand{\dtoprule}{\specialrule{\heavyrulewidth}{\abovetopsep}{1.2pt}\specialrule{\heavyrulewidth}{0pt}{\belowrulesep}}
\newcommand{\dbottomrule}{\specialrule{\heavyrulewidth}{\aboverulesep}{1.2pt}\specialrule{\heavyrulewidth}{0pt}{\belowbottomsep}}

\newcommand{\tdci}[3]{\makecell[r]{#1\\[-1pt]{\scriptsize[#2, #3]}}}

\begin{document}
\let\WriteBookmarks\relax
\renewcommand{\floatpagefraction}{0.85}
\renewcommand{\topfraction}{0.95}
\renewcommand{\bottomfraction}{0.6}
\renewcommand{\textfraction}{0.05}
\renewcommand{\dbltopfraction}{0.95}
\renewcommand{\dblfloatpagefraction}{0.85}
\setcounter{topnumber}{3}
\setcounter{dbltopnumber}{3}
\setcounter{totalnumber}{5}

\shorttitle{Agreement as a deferral signal in multi-agent medical
question answering}
\shortauthors{H.~Nguyen Van et~al.}

\title [mode = title]{Multi-agent LLM
consultation increases agreement but weakens it as a deferral signal
in medical question answering}
\tnotemark[1]
\tnotetext[1]{Code, prompts, the vote logs of every run and the analysis scripts that reproduce every table and figure are available at \url{https://github.com/phenikaa-a2i-lab/MACE}.}

\author[1]{Hoang {Nguyen Van}}[orcid=0009-0009-0756-6641]
\ead{23010101@st.phenikaa-uni.edu.vn}
\credit{Conceptualization, Methodology, Software, Investigation, Formal analysis, Data curation, Visualization, Writing -- original draft}

\author[1]{Trang {Kieu Thi Thu}}[orcid=0009-0000-5738-1711]
\ead{24100093@st.phenikaa-uni.edu.vn}
\credit{Software, Investigation, Validation, Data curation, Writing -- review \& editing}

\author[1]{Trang {Mai Xuan}}[orcid=0000-0002-9666-0198]
\ead{trang.maixuan@phenikaa-uni.edu.vn}
\credit{Methodology, Supervision, Writing -- review \& editing}

\author[2]{Lam {Pham Tien}}[orcid=0000-0002-9698-1688]
\ead{lam.phamtien@phenikaa-uni.edu.vn}
\credit{Methodology, Supervision, Writing -- review \& editing}

\author[1]{Dat {Nguyen Huu}}[orcid=0009-0005-1796-2198]
\cormark[1]
\ead{dat.nguyenhuu@phenikaa-uni.edu.vn}
\credit{Conceptualization, Supervision, Project administration, Resources, Writing -- review \& editing}

\affiliation[1]{organization={Applied AI Lab, Phenikaa School of Computing, Phenikaa University},
    city={Hanoi},
    country={Vietnam}}

\affiliation[2]{organization={Phenikaa School of Computing, Phenikaa University},
    city={Hanoi},
    country={Vietnam}}

\cortext[cor1]{Corresponding author}

\begin{abstract}
In medical question answering, agreement among independently sampled answers can help determine when to defer to a clinician.
Multi-agent consultation lets agents revise their answers after reading a report synthesised from their initial analyses, so the same report can influence multiple votes.
We tested how consultation changes the ability of vote agreement to identify incorrect answers.
For each of three open-weight models, we ran a nine-call consultation protocol on MedQA-USMLE and a 2,000-question subset of VM14K, a Vietnamese medical question-answering benchmark.
We recorded every vote before and after consultation.
Consultation increased unanimity in every model--benchmark setting and reduced the AUROC of vote agreement for detecting incorrect answers from 62--69 to 54--57, near the chance level of 50.
Dissenting agents tended to adopt the initial majority answer even when it was incorrect, and a simulation showed that this alone is enough to explain the loss.
On MedQA-USMLE, among answers that were unanimous after consultation, accuracy was 81--94\% when the initial votes were unanimous and 58--63\% when the initial votes were split 2--1.
In additional experiments on the VM14K subset, the decline in AUROC persisted in a council of three different models and in a variant in which agents read one another's full analyses.
These results support using agreement among pre-consultation votes to guide deferral.
\end{abstract}

\begin{highlights}
\item Consultation makes agents unanimous more often and unanimous answers less accurate
\item Agreement after consultation detects errors at near-chance level (AUROC 54--57)
\item The loss persists with three different models and without the synthesised report
\item Where agents disagree, consultation adds little to a majority vote over the analyses
\item Defer on the votes cast before interaction; consult only where samples disagree
\end{highlights}

\begin{keywords}
multi-agent systems \sep large language models \sep medical question answering \sep selective prediction \sep deferral \sep Vietnamese
\end{keywords}

\maketitle

\section{Introduction}
\label{sec:introduction}

Multi-agent systems for medical question answering bring together large language model (LLM) agents to analyse a question and reach a joint decision \cite{tang2024medagents,kim2024mdagents}.
A system intended to support clinicians must also indicate which answers warrant review \cite{geifman2017selective,savage2025uncertainty}.
For a single model, agreement among independently sampled answers can help distinguish correct from incorrect answers, including on medical questions \cite{xiong2024can,savage2025uncertainty}.
Agreement is also available in multi-agent systems that record individual votes.
Its interpretation changes, however, when agents read one another's reasoning, or a summary of it, before casting those votes.

Consider two medical questions for which all three agents agree after consultation (Fig.~\ref{fig:protocol}).
On one question, the agents were unanimous before consultation.
On the other, their initial votes were split two to one.
A deferral rule based only on the final vote count assigns both answers the same score.
Yet on MedQA-USMLE, final answers were correct in 81--94\% of cases in the first group and 58--63\% in the second.
The initial votes distinguish two groups of questions with substantially different error rates even though both end in unanimity.

Prior work documents several ways in which consultation can change an agent's answer.
Agents sometimes abandon a correct minority answer after reading their peers' reasoning \cite{wynn2025talk,yao2025peacemaker,pitre2025consensagent}, and an audit of medical multi-agent systems found that correct minority views can be suppressed and that information can be lost when the analyses are synthesised \cite{gu2025medagentaudit}.
Comparisons with majority voting and strong single-model baselines have also found inconsistent accuracy gains from debate, including on medical QA tasks \cite{choi2025debate,zhu2025medagentboard}.
Human groups show the same tension.
Pooling independent diagnoses can improve accuracy \cite{kurvers2016boosting,barnett2019comparative}, whereas seeing others' estimates narrows opinions without improving them \cite{lorenz2011social}.
For deferral, this raises a further question.
As agents converge, how much does their agreement still tell us about whether the final answer is correct?

Wang et al.~\cite{wang2026unanimously} examined this question in a multi-round medical retrieval system.
First-round agreement helped identify incorrect final answers, whereas terminal agreement performed near chance.
Their system stops once its agents agree, so terminal agreement is the same for every question that reaches unanimity, and part of its near-chance result follows from this design.
We study a protocol with a fixed stopping point.
Every question undergoes one round of consultation, and agents can still disagree at the end.
This lets us measure how the predictive value of agreement changes after a single exchange and trace which votes changed.

We adapted the shared-report consultation procedure of MedAgents \cite{tang2024medagents}.
Three agents with assigned roles independently analyse each question and cast an initial vote.
Each then reads a report synthesised from all three analyses and votes again.
Within each run, we evaluate both sets of votes against the correctness of the same final answer using AUROC and risk--coverage curves.
We evaluate three open-weight models on MedQA-USMLE \cite{jin2021medqa} and a 2,000-question subset of the Vietnamese benchmark VM14K \cite{nguyen2025vm14k}, with a separate replication on the full VM14K set.
We include five-sample self-consistency as an independent-sampling baseline \cite{wang2023selfconsistency}.

Our contributions are as follows.
\begin{itemize}
\item We measure how a single round of consultation changes the value of vote agreement for deferral. Consultation increased unanimity in every model--dataset setting, while the AUROC for detecting incorrect final answers fell from 62--69 to 54--57 on a 0--100 scale, where 50 is chance. In our AUROC comparison, post-consultation agreement ranked below all evaluated signals derived from independent reasoning (Sections~\ref{sec:res-inflation}, \ref{sec:res-signal} and~\ref{sec:res-league}). On the VM14K subset, the decline in AUROC persisted in two further experiments. One used a council of three different models, and the other replaced the shared report with the agents' full analyses (Section~\ref{sec:res-robust}).

\item We trace how agents revise their votes when the initial council is split two to one. When the dissenting agent held the only correct answer, it switched to the majority's incorrect answer in 59--72\% of cases. A simulation on the recorded votes shows that this indifference, not convergence, is enough to explain the loss (Section~\ref{sec:res-mechanism}).

\item We evaluate the initial votes for both deferral and early exit. For Qwen3-32B on MedQA-USMLE at 80\% coverage, the error rate among retained answers was 10.2\% when ranked by initial agreement and 15.5\% when ranked by final agreement. Stopping when the initial votes were unanimous yielded accuracy within 0.4 percentage points of the full protocol, with an average of 4.5--7.0 model calls per question instead of nine. We also evaluate a selective consultation policy, proposed to reduce cost~\cite{fan2026imad}, in which agreement among five independent samples decides which questions proceed to consultation (Sections~\ref{sec:res-signal} and~\ref{sec:res-remedies}).
\end{itemize}

We release the pre- and post-consultation vote logs used in the English and Vietnamese evaluations.

Section~\ref{sec:related} reviews related work, Section~\ref{sec:methods} describes the data, protocol and metrics, Section~\ref{sec:results} reports the results, Section~\ref{sec:discussion} discusses their implications, Section~\ref{sec:limitations} states the limitations and outlines future work, and Section~\ref{sec:conclusion} concludes.

\section{Related work}
\label{sec:related}

\subsection{Multi-agent debate and consultation}

In multi-agent debate, agents answer a question, exchange arguments and revise their answers \cite{du2024debate}.
Later systems differ in who takes part and in what the agents share.
ReConcile combines agents built on different models through confidence-weighted voting \cite{chen2024reconcile}.
MedAgents assigns specialist roles and has the specialists review a shared report synthesised from their analyses \cite{tang2024medagents}, and MDAgents uses the model's own estimate of question difficulty to choose between a single agent and a team \cite{kim2024mdagents}, an estimate that can itself serve as a confidence signal.
All of these designs assume that exchanging analyses improves the final answer.

Evaluations that control for cost have qualified this assumption.
Gains from debate depend on the task, the models and the baseline \cite{smit2024mad,wang2024rethinking}.
Choi et al.~\cite{choi2025debate} traced much of the gain to the majority vote over the initial answers and showed that the exchange itself does not raise the expected share of correct answers.
Medical studies report a similarly mixed picture when teams of agents are compared with strong single models \cite{zhu2025medagentboard}, although selecting an answer through peer review has improved accuracy on MedQA-USMLE \cite{zhan2026peer}.
These studies judge consultation by its accuracy.
They do not ask whether the agreement it produces still shows which answers are reliable.
Most of them also use English questions, whereas VM14K, a Vietnamese benchmark of 14,000 multiple-choice questions across 34 specialties, has so far been used to evaluate single models \cite{nguyen2025vm14k}.

\subsection{Agreement as a confidence signal}

A system that supports clinicians must also rank its answers by reliability, so that the least reliable ones can be referred for review.
Selective prediction judges a system together with the score it uses to decide which answers to keep \cite{geifman2017selective}, and learning to defer also accounts for how well the expert would handle each referred case \cite{madras2018predict,mozannar2020consistent}.
Both are only as good as that score.
For LLMs, a strong score comes from asking the same question several times.
Agreement among sampled reasoning paths, first used to choose an answer \cite{wang2023selfconsistency}, also works as a confidence score \cite{xiong2024can}.
On medical diagnosis and treatment tasks, this sample agreement separated correct from incorrect answers better than the confidence a model states or its token probabilities \cite{savage2025uncertainty}.
The score works because the samples are drawn independently, so that agreement reflects how consistently the model reaches an answer.
The votes of a multi-agent council meet this condition only in part.
Agents vote separately at first, but they receive different role prompts and can share errors even when they are built on different models \cite{kim2025correlated}, and after consultation they have all read the same material.

\subsection{Conformity and abstention under interaction}

Interaction can change answers in ways that weaken agreement as evidence.
Agents change their answers to match their peers \cite{wynn2025talk,yao2025peacemaker,pitre2025consensagent}.
Conformity grows with the size of the majority and the length of the interaction \cite{weng2025conformity}, and agents with similar beliefs can reinforce one another into echo chambers \cite{estornell2024multi}.
An audit of six medical multi-agent frameworks traced flawed consensus to weaknesses shared by all agents and found that correct minority opinions were suppressed during discussion \cite{gu2025medagentaudit}.
Agents can also reach the same answer for conflicting reasons, so that agreement on the answer overstates agreement on the reasoning \cite{wang2026illusion}.
These studies show that consensus can form without better answers, but they describe the answers rather than the reliability signal that the consensus provides.

Two recent methods address that signal by adding components so that interacting agents can abstain.
Wang et al.~\cite{wang2026unanimously} built an abstention rule from the voting history of a multi-round medical system and tested whether its answer survived evidence retrieved against it.
Another study combined the probabilities reported by different agents into conformal prediction sets and sent cases with several options to human review \cite{wang2026debate}.
Both methods add new parts to the system.
Neither asks how much of the signal a single exchange removes when agents can keep their votes, or whether the signal survives in the votes that the protocol already records.

\section{Materials and methods}
\label{sec:methods}

\subsection{Datasets}
\label{sec:data}

MedQA-USMLE \cite{jin2021medqa} consists of four-option questions in English from the United States Medical Licensing Examination.
We used its development split of 1,272 questions.
VM14K \cite{nguyen2025vm14k} is a Vietnamese benchmark of multiple-choice medical questions labelled by specialty and difficulty.
We kept its 11,094 four-option questions and drew a random subset of 2,000 for the main experiments.
The subset matches the full set to within 0.5 percentage points in the share of any difficulty level and 1.3 points in the share of any specialty, and its counts are given in \ref{app:config}.
Unless stated otherwise, VM14K refers to this subset.

\subsection{Models and configurations}
\label{sec:models}\label{sec:config}

We evaluated three open-weight instruction-tuned models, Qwen3-32B \cite{yang2025qwen3}, Llama-3.3-70B and Llama-3.1-8B \cite{grattafiori2024llama3}, served locally with the reasoning mode of Qwen3 disabled.
All runs on MedQA-USMLE and the VM14K subset, which form the primary analysis, used one setup that we call the unified configuration, with 4-bit AWQ checkpoints \cite{lin2024awq}, a single inference engine, an output limit of 1,024 tokens per call and full logging of raw outputs.
Every strategy was run with three seeds, except the runs on the VM14K subset, which used two.
The full VM14K set was run under a legacy configuration with lower output limits and partial logging, and we report it as a replication in \ref{app:config}.

\subsection{Reasoning strategies}
\label{sec:strategies}

Four strategies form the main comparison.
Zero-shot prompting asks for the option letter only, and chain-of-thought (CoT) prompting asks for step-by-step reasoning that ends with the letter \cite{wei2022cot}.
Self-consistency (CoT-SC) draws five CoT samples at a temperature of 0.5, as in the original method, and returns the plurality answer \cite{wang2023selfconsistency}.
All other calls use a temperature of 0.3.
Consultation follows the five-step design of MedAgents \cite{tang2024medagents} and uses nine calls per question, so CoT-SC, with five calls, is the single-model reference closest to it in cost.
Four further strategies were run for the confidence signals they give, which Section~\ref{sec:res-league} compares.
They are Self-Refine \cite{madaan2023selfrefine}, voting over three zero-shot runs with shuffled options, MDAgents \cite{kim2024mdagents} and, on the full VM14K set, a consultation variant with roles taken from the specialty labels, and their accuracy is reported in \ref{app:results}.

\subsection{Consultation protocol}
\label{sec:protocol}

\begin{figure*}[pos=tbp]
\centering
\includegraphics[width=\linewidth]{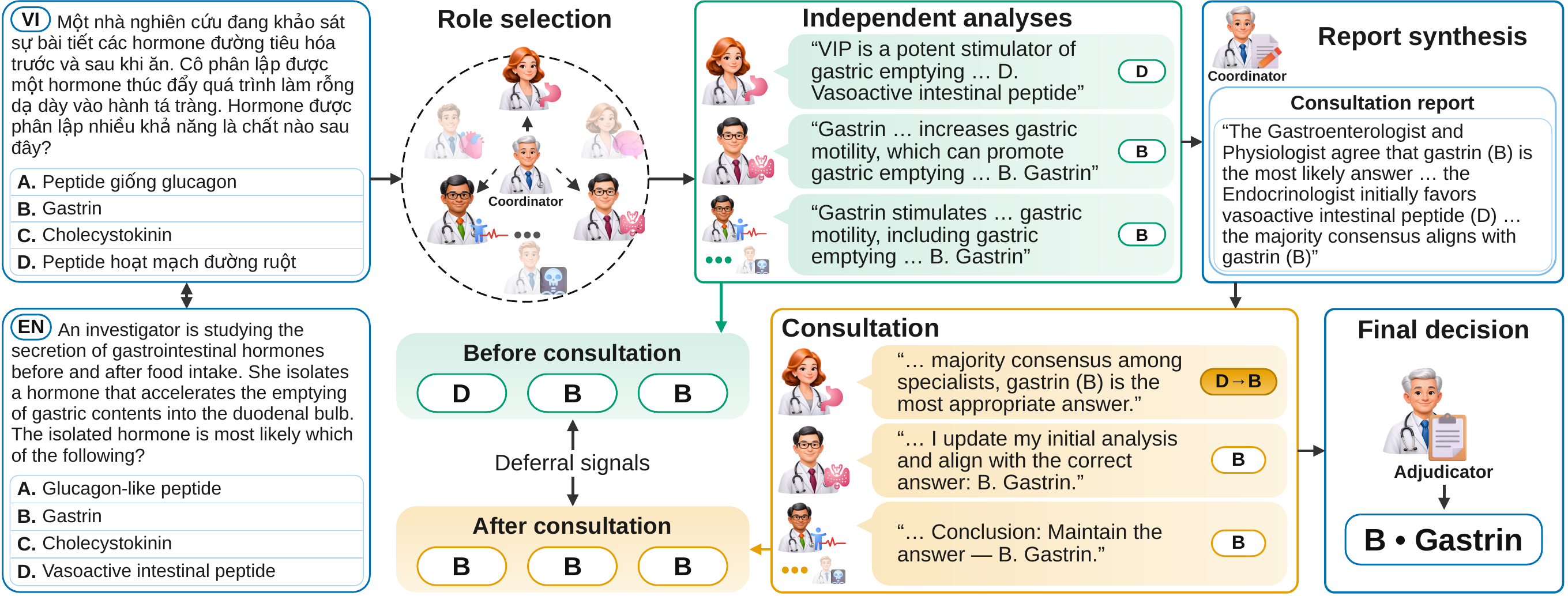}
\caption{The consultation protocol. Three role-conditioned agents analyse the question independently (step~2), a report is synthesised from their analyses (step~3), each agent revises its answer after reading it (step~4) and a decision is made (step~5). Agreement among the three votes is read before (blue) and after (orange) consultation and evaluated as a deferral signal.}
\label{fig:protocol}
\end{figure*}

Fig.~\ref{fig:protocol} shows the consultation protocol, our single-round simplification of the published framework.
In step~1 the model names three specialist roles relevant to the question.
In step~2 each role, in a separate call and without access to the other roles, analyses the question and each option and states an answer.
In step~3 a coordinator call summarises the analyses into a consultation report of three to five sentences.
In step~4 each role receives the question, the report and its own analysis, and states whether it maintains or updates its answer.
In step~5 a decision call receives the three revised positions and outputs one option letter, which is the answer of the system.
We call the answers stated in step~2 the \emph{pre-consultation votes} and those stated in step~4 the \emph{post-consultation votes}.

Two properties of this implementation matter for interpreting the results.
Each step passes on only the opening of each analysis or revision, cut at the last complete sentence within 500 characters.
Steps~3 and~5 also receive the answer extracted from each agent.
In step~4 an agent sees the report and the opening of its own analysis, so the report is the only channel between agents.

To test whether the results depend on using three copies of one model and on the report being the only channel between agents, we ran two variants on the VM14K subset.
In the \emph{heterogeneous council}, the three roles are played by Qwen3-32B, Llama-3.3-70B and Llama-3.1-8B in fixed order, while role selection, the report and the decision remain with Qwen3-32B and all other settings are unchanged.
In \emph{consultation without report}, step~3 is omitted and each agent in step~4 receives the full analyses, each labelled with its role and answer, in place of the report, so the protocol uses eight calls.
The agents are three instances of Qwen3-32B, as in the standard protocol.

\subsection{Vote extraction and evaluation metrics}
\label{sec:votes}\label{sec:metrics}

Every agreement measure in this study is computed from the option letters that the agents state.
A rule-based parser extracted these letters at run time.
Where it found none, we used the last explicit, non-negated answer in the stored raw output.
The two methods agreed in 93--100\% of the cases where both found a vote, and fewer than 0.4\% of analyses and 1.9\% of revisions remained without a vote.
A missing vote is counted as dissenting, which lowers measured agreement.
Steps that had used a wrongly parsed vote were re-executed, and all results come from these re-executed runs, as detailed in \ref{app:config}; the run-time votes alone give the same findings (Table~\ref{tab:sens}a).

For a set of votes, the agreement is the share cast for the plurality answer, and the votes are unanimous when the agreement is one.
We use agreement as a deferral signal, a score by which a system ranks its answers for deferral.
Pre- and post-consultation agreement, each over three votes, rank the final answer of the consultation protocol.
Agreement among the $k=5$ CoT-SC samples, or among the first $k=3$ of them, ranks the plurality answer of those samples.

We evaluate each signal by selective prediction \cite{geifman2017selective}.
Questions are answered in decreasing order of the signal and the rest are deferred.
The selective risk, which is the error rate among answered questions, is traced against the coverage, which is the share of questions answered.
We report the area under this risk--coverage curve (AURC), where lower is better, and the area under the ROC curve (AUROC) of the signal for detecting an incorrect final answer, both on a 0--100 scale, where an AUROC of 50 is chance.
Agreement among three votes takes only three values, so ties are frequent and limit how finely the signal can rank answers.
Tied pairs count one half in the AUROC, and the AURC is averaged over random orderings of tied questions.
CoT-SC with $k=3$ gives a reference signal with the same three values.
The share of unanimous votes and the accuracy of the unanimous answers, both in Table~\ref{tab:agreement}, describe the largest tied group.
Section~\ref{sec:res-league} compares only signals that the strategies produce while answering, so verbalised confidence and token probabilities are not included.

Four further quantities describe how the votes change during consultation.
The initial-majority accuracy is the accuracy of the plurality of the pre-consultation votes, and the oracle accuracy is the share of questions on which at least one pre-consultation vote is correct.
A question is unanimous and wrong when all three votes agree on an incorrect option.
The minority override rate is the share of questions with exactly one correct pre-consultation vote, whether the split is two to one or three ways, on which the final answer is incorrect.
All metrics are computed per seed and averaged over seeds.

\subsection{Statistical analysis}
\label{sec:stats}

All confidence intervals are 95\% nonparametric bootstrap intervals over questions.
For accuracy we resample the per-question mean over seeds, differences between strategies are paired by question, and intervals for differences in AURC and AUROC resample questions within each seed and pool the differences across seeds.
Intervals are not adjusted for multiple comparisons.
For the two pre-specified families of six comparisons each, consultation against CoT-SC and post- against pre-consultation AUROC, two-sided bootstrap $p$-values are Holm-adjusted within each family and reported alongside Tables~\ref{tab:accuracy} and~\ref{tab:selective}.
The numbers of resamples are given in \ref{app:config}.

\section{Results}
\label{sec:results}

We report results for MedQA-USMLE and the VM14K subset.
The replication on the full VM14K set is summarised in Section~\ref{sec:res-robust} and given in full in \ref{app:config}.
On the subset we ran CoT-SC and the consultation protocol for the three models and the two protocol variants for Qwen3-32B, each with two seeds.
Between seeds, unanimity varied by at most 3.1 points, accuracy by at most 2.0 points and the AUROC of agreement by at most 3.6 points in any cell.

\subsection{Accuracy}
\label{sec:res-accuracy}

Consultation did not improve on self-consistency (Table~\ref{tab:accuracy}).
On MedQA-USMLE it was less accurate than CoT-SC for all three models, by 0.92 to 4.38 percentage points, and the difference remained significant after Holm adjustment for Qwen3-32B and Llama-3.1-8B.
It also used 1.4 to 1.7 times as many tokens per question.
On VM14K the two strategies did not differ for any model, and consultation used 1.8 to 2.1 times the tokens.
Consultation also added little to a simple vote over its own analyses.
In every cell, the accuracy of the plurality of the three pre-consultation votes was within 1.2 points of that of the final answer, and within 0.8 points on MedQA-USMLE (Table~\ref{tab:agreement}).

\begin{table*}[pos=tbp]\centering\small
\begin{threeparttable}
\caption{Accuracy of the four main strategies.}\label{tab:accuracy}
\setlength{\tabcolsep}{4pt}
\begin{tabular}{@{}llrrrrrr@{}}\dtoprule
\multicolumn{2}{@{}l}{} & \multicolumn{3}{c}{MedQA-USMLE} & \multicolumn{3}{c}{VM14K} \\
\cmidrule(lr){3-5}\cmidrule(l){6-8}
Strategy & Calls & Qwen3-32B & Llama-3.3-70B & Llama-3.1-8B & Qwen3-32B & Llama-3.3-70B & Llama-3.1-8B \\ \midrule
Zero-shot & 1 & \tdci{73.8}{71.3}{76.2} & \tdci{86.8}{85.0}{88.6} & \tdci{58.3}{55.9}{60.9} & -- & -- & -- \\
CoT & 1 & \tdci{82.9}{81.2}{84.7} & \tdci{\underline{90.9}}{89.4}{92.2} & \tdci{66.8}{64.6}{68.9} & -- & -- & -- \\
CoT-SC & 5 & \tdci{\textbf{86.3}}{84.6}{88.1} & \tdci{\textbf{91.7}}{90.3}{93.0} & \tdci{\textbf{71.5}}{69.3}{73.6} & \tdci{\textbf{71.9}}{70.0}{73.6} & \tdci{69.3}{67.3}{71.3} & \tdci{\textbf{49.7}}{47.6}{51.7} \\
Consultation & 9 & \tdci{\underline{83.3}}{81.3}{85.2} & \tdci{90.7}{89.3}{92.2} & \tdci{\underline{67.1}}{65.0}{69.2} & \tdci{71.1}{69.2}{72.8} & \tdci{\textbf{69.4}}{67.5}{71.3} & \tdci{49.1}{47.2}{51.2} \\
\midrule
Consultation $-$ CoT-SC (pp) &  & \tdci{$-$3.09}{$-$4.25}{$-$1.99} & \tdci{$-$0.92}{$-$1.78}{$+$0.08} & \tdci{$-$4.38}{$-$6.16}{$-$2.54} & \tdci{$-$0.78}{$-$2.10}{$+$0.50} & \tdci{$+$0.10}{$-$1.20}{$+$1.38} & \tdci{$-$0.55}{$-$2.40}{$+$1.45} \\
\quad Holm-adjusted $p$ &  & $<$0.001 & 0.256 & $<$0.001 & 0.789 & 1.000 & 1.000 \\
 
\dbottomrule\end{tabular}
\begin{tablenotes}[flushleft]\footnotesize
\item[] VM14K, 2,000-question subset. Mean over seeds (three on MedQA-USMLE, two on VM14K) with the 95\% bootstrap CI in brackets below each value. --, not run on VM14K. Bold, most accurate strategy in each column. Underlined, second most accurate where four strategies are compared. Calls, model calls per question. pp, paired difference in percentage points, with Holm-adjusted $p$ from 10,000 paired resamples. Mean tokens per question, input and output, in the order of the model columns, were 4,452, 4,011 and 3,421 for CoT-SC and 6,638, 5,638 and 5,738 for consultation on MedQA-USMLE, and 2,606, 2,513 and 2,188 for CoT-SC and 4,689, 4,420 and 4,690 for consultation on VM14K.
\end{tablenotes}
\end{threeparttable}
\end{table*}

\subsection{Agreement before and after consultation}
\label{sec:res-inflation}

\begin{figure*}[pos=tbp]
\centering
\includegraphics[width=\linewidth]{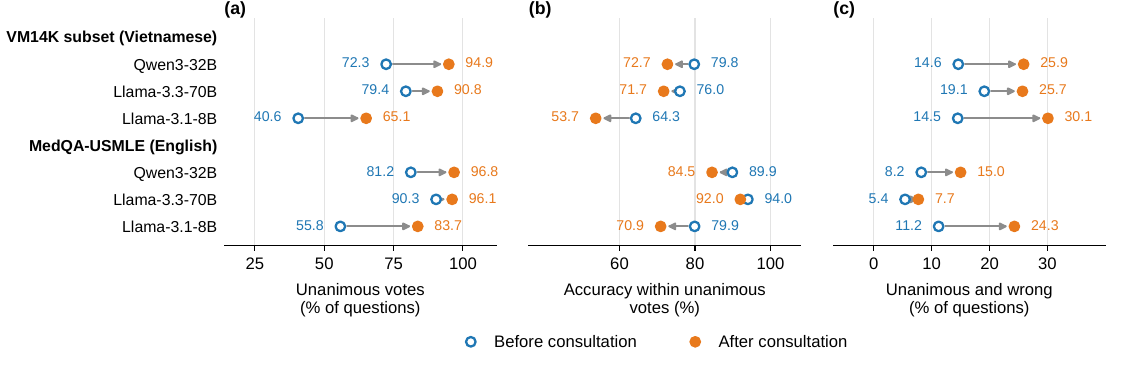}
\caption{Consultation inflates agreement without improving its reliability: (a) share of unanimous votes, (b) accuracy of the unanimous answer and (c) share of questions unanimous and wrong, before (open) and after (filled) consultation. VM14K: 2,000-question subset.}
\label{fig:inflation}
\end{figure*}

Consultation made the agents unanimous on more questions, and it made unanimity less reliable (Fig.~\ref{fig:inflation}, Table~\ref{tab:agreement}).
Unanimity rose in every cell, from 56--90\% to 84--97\% of questions on MedQA-USMLE and from 41--79\% to 65--95\% on VM14K.
Over the same step, the accuracy of the unanimous answer fell by 2.0--9.0 points on MedQA-USMLE and by 4.3--10.6 points on VM14K.
The share of questions on which the agents were unanimous and wrong rose in every cell, from 5--19\% to 8--30\% of all questions.

\begin{table*}[pos=tbp]\centering\small
\begin{threeparttable}
\caption{Agreement before and after consultation.}\label{tab:agreement}
\setlength{\tabcolsep}{3pt}
\begin{tabular}{@{}>{\raggedright\arraybackslash}p{29mm}lrrrrrr@{}}\dtoprule
\multicolumn{2}{@{}l}{} & \multicolumn{3}{c}{MedQA-USMLE} & \multicolumn{3}{c}{VM14K} \\
\cmidrule(lr){3-5}\cmidrule(l){6-8}
\multicolumn{2}{@{}l}{Measure} & Qwen3-32B & Llama-3.3-70B & Llama-3.1-8B & Qwen3-32B & Llama-3.3-70B & Llama-3.1-8B \\ \midrule
\multirow[t]{2}{=}{Unanimous votes} & Before & 81.2 & 90.3 & 55.8 & 72.3 & 79.4 & 40.6 \\
 & After & 96.8 & 96.1 & 83.7 & 94.9 & 90.8 & 65.1 \\
\midrule
\multirow[t]{2}{=}{Accuracy of unanimous answer} & Before & 89.9 & 94.0 & 79.9 & 79.8 & 76.0 & 64.3 \\
 & After & 84.5 & 92.0 & 70.9 & 72.7 & 71.7 & 53.7 \\
\midrule
\multirow[t]{2}{=}{Unanimous and wrong} & Before & 8.2 & 5.4 & 11.2 & 14.6 & 19.1 & 14.5 \\
 & After & 15.0 & 7.7 & 24.3 & 25.9 & 25.7 & 30.1 \\
\midrule
\multirow[t]{3}{=}{Accuracy} & Initial majority & 84.1 & 90.7 & 66.5 & 70.5 & 69.3 & 47.9 \\
 & Final answer & 83.3 & 90.7 & 67.1 & 71.1 & 69.4 & 49.1 \\
 & Oracle & 89.6 & 93.8 & 80.9 & 79.7 & 76.6 & 68.5 \\
 
\dbottomrule\end{tabular}
\begin{tablenotes}[flushleft]\footnotesize
\item[] VM14K, 2,000-question subset. Unanimous votes and unanimous and wrong are shares of all questions. Accuracy of unanimous answer, accuracy of the unanimous vote itself, read at step~2 before consultation and at step~4 after it. Initial majority, plurality of the pre-consultation votes. Final answer, answer of the consultation protocol. Oracle, share of questions with at least one correct pre-consultation vote.
\end{tablenotes}
\end{threeparttable}
\end{table*}

\subsection{Agreement as a deferral signal}
\label{sec:res-signal}

\begin{figure*}[pos=tbp]
\centering
\includegraphics[width=\linewidth]{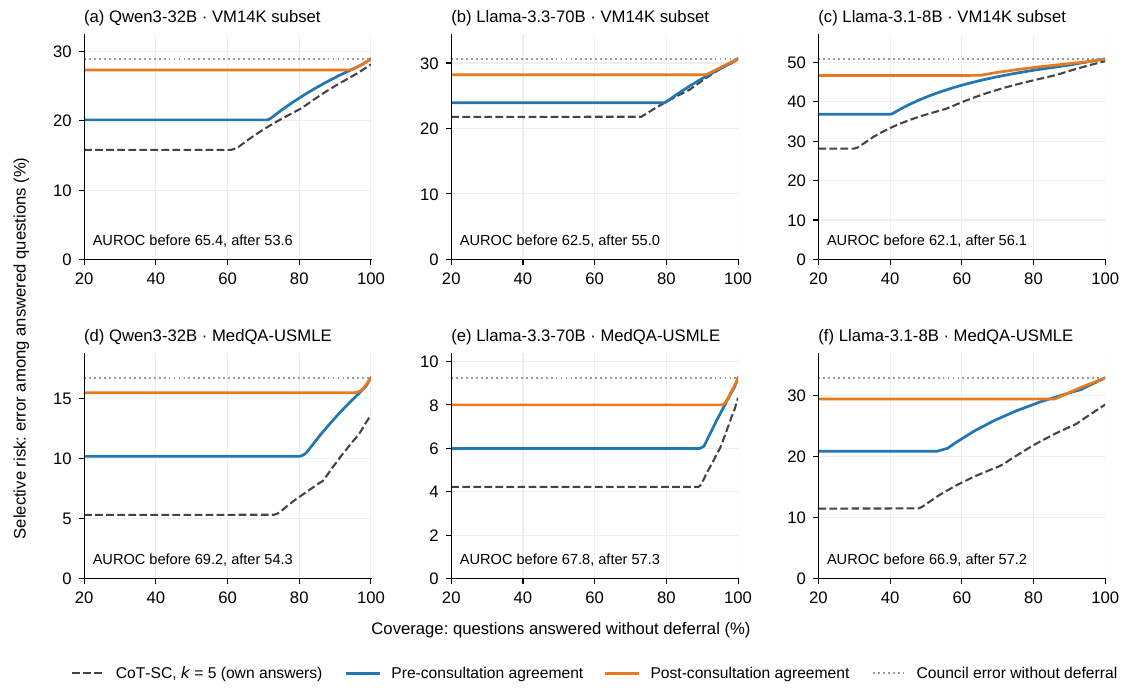}
\caption{Risk--coverage curves of agreement-based deferral: selective risk among answered questions as coverage varies, ties ordered at random. Pre- and post-consultation agreement rank the final answer of the protocol; CoT-SC ranks its own answers; the dotted line is the error rate without deferral. VM14K: 2,000-question subset.}
\label{fig:rc}
\end{figure*}

As a detector of incorrect final answers, agreement was much weaker after consultation than before it (Table~\ref{tab:selective}, Fig.~\ref{fig:rc}).
The AUROC fell from 66.9--69.2 to 54.3--57.3 on MedQA-USMLE and from 62.1--65.4 to 53.6--56.1 on VM14K, and every interval for the change excluded zero.
After consultation, the AURC was only 1.2--3.7 points below the error rate without deferral, which is the value an uninformative signal would reach.
The order of the signals was the same in every cell.
Agreement among five CoT-SC samples was the most informative, pre-consultation agreement came next, and post-consultation agreement was the least informative.
This weak ranking holds even though post-consultation agreement is closely tied to the final answer.
The decision call returned the plurality of the post-consultation votes on 90--99\% of questions in every cell, so post-consultation agreement was in effect the margin by which the final answer won its own vote.

The loss is also visible at fixed operating points.
At 80\% coverage, deferral by post-consultation agreement lowered the selective risk by only 1.2--3.8 points relative to the error rate without deferral.
For Qwen3-32B on MedQA-USMLE, the selective risk at this coverage was 10.2\% with pre-consultation agreement and 15.5\% with post-consultation agreement, against 16.7\% without deferral.
In five of the six cells, the votes were unanimous after consultation on more than 80\% of questions, so this operating point fell inside the tied group of unanimous questions.
The signal could not choose which of these questions to keep, and the expected risk equalled the error rate of the unanimous answers.
A simple rule that defers every question on which the agents are not unanimous after consultation would defer 3--35\% of questions and lower the error rate among the rest by only 1.2--4.6 points.
The same rule applied to the pre-consultation votes would defer 10--59\% of questions and lower the error rate by 3.3--15.2 points.
The loss did not depend on vote re-extraction, and it remained when only run-time votes were used (Table~\ref{tab:sens}a).

\begin{table*}[pos=tbp]\centering\small
\begin{threeparttable}
\caption{Agreement as a signal for detecting an incorrect answer.}\label{tab:selective}
\setlength{\tabcolsep}{2.5pt}
\begin{tabular}{@{}>{\raggedright\arraybackslash}p{13mm}lrrrrrr@{}}\dtoprule
\multicolumn{2}{@{}l}{} & \multicolumn{3}{c}{MedQA-USMLE} & \multicolumn{3}{c}{VM14K} \\
\cmidrule(lr){3-5}\cmidrule(l){6-8}
Metric & Signal & Qwen3-32B & Llama-3.3-70B & Llama-3.1-8B & Qwen3-32B & Llama-3.3-70B & Llama-3.1-8B \\ \midrule
\multirow[t]{6}{=}{AUROC ($\uparrow$)} & CoT-SC agreement, $k=5$ & \textbf{78.1} & \textbf{74.4} & \textbf{74.1} & \textbf{70.2} & \textbf{65.9} & \textbf{67.9} \\
 & CoT-SC agreement, $k=3$ & \underline{74.0} & \underline{68.7} & \underline{70.7} & \underline{67.6} & \underline{62.8} & \underline{65.6} \\
 & Pre-consultation agreement & 69.2 & 67.8 & 66.9 & 65.4 & 62.5 & 62.1 \\
 & Post-consultation agreement & 54.3 & 57.3 & 57.2 & 53.6 & 55.0 & 56.1 \\
\cmidrule(l){2-8}
 & Post $-$ pre & \tdci{$-$15.0}{$-$18.7}{$-$11.4} & \tdci{$-$10.6}{$-$15.0}{$-$6.1} & \tdci{$-$9.7}{$-$13.1}{$-$6.2} & \tdci{$-$11.8}{$-$14.1}{$-$9.4} & \tdci{$-$7.5}{$-$9.5}{$-$5.6} & \tdci{$-$6.0}{$-$8.7}{$-$3.4} \\
\cmidrule(l){2-8}
 & Pre-consultation, own answer\tnote{a} & 67.6 & 67.7 & 67.6 & 66.4 & 62.5 & 64.9 \\
\midrule
\multirow[t]{6}{=}{AURC ($\downarrow$)} & CoT-SC agreement, $k=5$ & \textbf{6.2} & \textbf{4.4} & \textbf{15.9} & \textbf{18.3} & \textbf{22.8} & \textbf{36.9} \\
 & CoT-SC agreement, $k=3$ & \underline{7.8} & \underline{5.4} & \underline{19.1} & \underline{20.9} & \underline{24.4} & \underline{40.5} \\
 & Pre-consultation agreement & 10.9 & 6.0 & 23.3 & 21.8 & \underline{24.4} & 41.8 \\
 & Post-consultation agreement & 15.5 & 7.8 & 29.4 & 27.6 & 28.2 & 47.2 \\
\cmidrule(l){2-8}
 & Post $-$ pre & \tdci{$+$4.7}{$+$3.3}{$+$6.2} & \tdci{$+$1.9}{$+$1.0}{$+$3.0} & \tdci{$+$6.1}{$+$3.8}{$+$8.4} & \tdci{$+$5.8}{$+$4.4}{$+$7.2} & \tdci{$+$3.7}{$+$2.6}{$+$4.9} & \tdci{$+$5.3}{$+$3.4}{$+$7.2} \\
\cmidrule(l){2-8}
 & Error rate without deferral & 16.7 & 9.3 & 32.9 & 28.9 & 30.6 & 50.9 \\
 
\dbottomrule\end{tabular}
\begin{tablenotes}[flushleft]\footnotesize
\item[] VM14K, 2,000-question subset. $\uparrow$, higher is better. $\downarrow$, lower is better. CoT-SC agreement ranks the CoT-SC answer, and pre- and post-consultation agreement rank the final answer of the consultation protocol. Bold, best signal in each column. Underlined, second best. Differences with 95\% bootstrap CI in brackets. All AUROC differences have Holm-adjusted $p<0.001$. The error rate without deferral is that of the final answer of the consultation protocol, and it equals the expected AURC of an uninformative signal for that answer.
\item[a] Ranks the plurality answer of the pre-consultation votes rather than the final answer, so it is not marked.
\end{tablenotes}
\end{threeparttable}
\end{table*}

\subsection{Why the signal is lost}
\label{sec:res-saturation}\label{sec:res-mechanism}

A signal that saturates loses some discrimination.
Three analyses show that the loss came from how consultation reached unanimity, not from saturation alone.

First, a unanimous post-consultation vote pooled questions of very different reliability (Table~\ref{tab:strata}).
On MedQA-USMLE, the final answer was correct on 81.0--94.1\% of the questions on which the agents had already been unanimous before consultation.
It was correct on 57.5--62.7\% of the questions that had been split two to one and, for Qwen3-32B and Llama-3.1-8B, on 39--40\% of those on which all three agents had disagreed.
VM14K showed the same pattern, with 66.5--79.7\%, 41.9--50.8\% and 30.0--39.4\%.
The last two groups made up 7--39\% of the unanimous post-consultation votes on MedQA-USMLE and 15--51\% on VM14K.
On the two-to-one questions that became unanimous, accuracy was within $-$1.4 to $+$5.5 points of the accuracy of the initial majority on all two-to-one questions (Tables~\ref{tab:strata} and~\ref{tab:split}), so consultation promoted questions to unanimity whether or not the majority was correct.
Within the unanimous post-consultation votes, pre-consultation agreement still separated correct from incorrect answers, with an AUROC of 62.8--67.3 on MedQA-USMLE and 59.0--63.8 on VM14K.
The information was present before consultation but could no longer be read from the votes afterwards.

Second, unanimity reached through consultation was less stable.
When a question was run again with another seed, the answer of a unanimous council changed in 2.5--18.6\% of cases on MedQA-USMLE and in 6.0--28.1\% on VM14K.
For three unanimous pre-consultation votes, which come from the same calls at the same temperature, the rates were 0.7--7.1\% and 2.4--13.2\%.
Unanimity after consultation changed two to four times as often as unanimity among the same three votes before it.
Part of this gap comes from the questions that consultation added to the unanimous group, which were answered correctly less often, so the comparison does not show how much consultation changed questions that were already unanimous.

Third, consultation did not resolve the disagreements it was meant to resolve (Table~\ref{tab:split}).
On questions split two to one, the final answer was between 3.5 points less and 1.6 points more accurate than the initial majority.
The only interval that excluded zero was for Qwen3-32B on MedQA-USMLE, where consultation was worse.
A correct answer was present among the three analyses in 69--93\% of these questions.

These patterns follow from how the votes moved (Table~\ref{tab:mechanism}, Fig.~\ref{fig:A1}).
On questions split two to one, the minority agent adopted the majority answer in 59--72\% of cases when its own answer was correct and in 63--78\% when it was wrong.
When exactly one agent held the correct answer before consultation, the final answer was wrong in 65--78\% of cases.
The accuracy of the individual votes changed by three points or less on MedQA-USMLE.
Consultation moved votes towards the majority without making them more accurate.

A simulation on the recorded votes shows that this indifference is enough to explain the loss (Table~\ref{tab:null}).
We regenerated the post-consultation votes of the two-to-one questions 200 times under three rules and kept all other votes as observed.
With the observed switching rates, the simulation reproduced the observed AUROC to within 1.6 points.
When every minority agent switched at one pooled rate, whether right or wrong, the AUROC was 54.4--56.9.
When only wrong minority agents switched, at the same level of unanimity, it was 61.0--68.1, close to its value before consultation.
Convergence alone therefore does not remove the signal; yielding regardless of correctness does.

The same holds when agents see every answer in full.
In the variant without report, a correct minority agent still adopted the majority answer in about two thirds of cases.
The simulation gave 54.2 under the pooled rule and 63.8 when only wrong agents switched, against 55.3 observed.

\begin{table*}[pos=tbp]\centering\small
\begin{threeparttable}
\caption{Composition and stability of unanimous post-consultation votes.}\label{tab:strata}
\setlength{\tabcolsep}{2.5pt}
\begin{tabular}{@{}>{\raggedright\arraybackslash}p{22mm}lrrrrrr@{}}\dtoprule
\multicolumn{2}{@{}l}{} & \multicolumn{3}{c}{MedQA-USMLE} & \multicolumn{3}{c}{VM14K} \\
\cmidrule(lr){3-5}\cmidrule(l){6-8}
\multicolumn{2}{@{}l}{Measure} & Qwen3-32B & Llama-3.3-70B & Llama-3.1-8B & Qwen3-32B & Llama-3.3-70B & Llama-3.1-8B \\ \midrule
\multicolumn{2}{@{}l}{Not unanimous before (\%)} & 16.5 & 6.7 & 39.2 & 23.8 & 14.5 & 51.0 \\
\midrule
\multirow[t]{3}{=}{Final accuracy by prior split (\%)} & 3--0 & 89.8 & 94.1 & 81.0 & 79.7 & 76.0 & 66.5 \\
 & 2--1 & 59.1 & 62.7 & 57.5 & 50.8 & 47.4 & 41.9 \\
 & 1--1--1 & 39.5 & 87.5\tnote{a} & 38.7 & 39.4 & 30.0 & 36.9 \\
\midrule
\multirow[t]{4}{=}{Answer changes on re-run (\%, $\downarrow$)} & CoT, $k=3$ & 2.4 & 1.2 & \underline{6.0} & 4.4 & 2.7 & 13.3 \\
 & Pre-consultation votes & \underline{1.7} & \underline{0.7} & 7.1 & \underline{3.2} & \underline{2.4} & \underline{13.2} \\
 & Council & 6.7 & 2.5 & 18.6 & 11.7 & 6.0 & 28.1 \\
 & CoT-SC, $k=5$ & \textbf{0.6} & \textbf{0.2} & \textbf{2.0} & \textbf{1.1} & \textbf{0.9} & \textbf{4.7} \\
\midrule
\multicolumn{2}{@{}l}{Pre-agreement AUROC} & 67.3 & 62.8 & 64.5 & 63.8 & 59.0 & 62.6 \\
 
\dbottomrule\end{tabular}
\begin{tablenotes}[flushleft]\footnotesize
\item[] VM14K, 2,000-question subset. Not unanimous before, share of unanimous post-consultation votes that were not unanimous before consultation. Final accuracy, accuracy of the step-5 answer within the unanimous post-consultation votes, grouped by the vote split at step~2, so its 3--0 row differs slightly from the accuracy of the unanimous answer before consultation in Table~\ref{tab:agreement}. Answer changes on re-run, share of unanimous outcomes whose answer changes when the question is run with another seed, for three CoT samples, the three pre-consultation votes, the council after consultation and five CoT-SC samples. The share is taken over ordered pairs of seeds in which the outcome was unanimous in the first seed, and a change is counted when the answer from the same source in the second seed differs, whether or not that answer is unanimous. The pre-consultation votes come from the same calls and the same temperature as the council. Bold, most stable in each column. Underlined, second most stable. Pre-agreement AUROC, AUROC of pre-consultation agreement for the final answer within the unanimous post-consultation votes.
\item[a] Fewer than ten questions per seed.
\end{tablenotes}
\end{threeparttable}
\end{table*}

\begin{table*}[pos=tbp]\centering\small
\begin{threeparttable}
\caption{Questions split two to one before consultation.}\label{tab:split}
\setlength{\tabcolsep}{3pt}
\begin{tabular}{@{}>{\raggedright\arraybackslash}p{24mm}lrrrrrr@{}}\dtoprule
\multicolumn{2}{@{}l}{} & \multicolumn{3}{c}{MedQA-USMLE} & \multicolumn{3}{c}{VM14K} \\
\cmidrule(lr){3-5}\cmidrule(l){6-8}
\multicolumn{2}{@{}l}{Measure} & Qwen3-32B & Llama-3.3-70B & Llama-3.1-8B & Qwen3-32B & Llama-3.3-70B & Llama-3.1-8B \\ \midrule
\multicolumn{2}{@{}l}{Questions per seed} & 220 & 115 & 476 & 500 & 372 & 971 \\
\midrule
\multirow[t]{3}{=}{Accuracy (\%)} & Initial majority & \underline{60.5} & \underline{61.2} & 52.0 & 48.1 & \underline{44.6} & 39.5 \\
 & Final answer & 57.1 & 60.9 & \underline{53.5} & \underline{49.7} & 44.4 & \underline{40.3} \\
 & CoT-SC & \textbf{69.5} & \textbf{68.4} & \textbf{60.5} & \textbf{55.1} & \textbf{46.2} & \textbf{42.6} \\
\midrule
\multicolumn{2}{@{}l}{Oracle (\%)} & 88.2 & 92.5 & 80.9 & 79.0 & 78.0 & 69.1 \\
\midrule
\multirow[t]{2}{=}{Difference (pp)} & Final $-$ majority & \tdci{$-$3.5}{$-$6.4}{$-$0.6} & \tdci{$-$0.3}{$-$3.8}{$+$3.5} & \tdci{$+$1.5}{$-$0.6}{$+$3.7} & \tdci{$+$1.6}{$-$0.6}{$+$3.6} & \tdci{$-$0.1}{$-$2.4}{$+$2.3} & \tdci{$+$0.8}{$-$1.2}{$+$2.8} \\
 & Final $-$ CoT-SC & \tdci{$-$12.4}{$-$16.2}{$-$8.5} & \tdci{$-$7.5}{$-$13.3}{$-$1.7} & \tdci{$-$6.9}{$-$9.7}{$-$3.9} & \tdci{$-$5.5}{$-$8.9}{$-$2.3} & \tdci{$-$1.7}{$-$5.5}{$+$2.3} & \tdci{$-$2.3}{$-$4.7}{$+$0.3} \\
 
\dbottomrule\end{tabular}
\begin{tablenotes}[flushleft]\footnotesize
\item[] VM14K, 2,000-question subset. Initial majority, plurality of the pre-consultation votes. Oracle, share of questions with at least one correct pre-consultation vote. Bold, most accurate of the initial majority, the final answer and CoT-SC in each column. Underlined, second most accurate. Differences are paired by question, in percentage points, with 95\% bootstrap CI in brackets. The same quantities for the two protocol variants are in Table~\ref{tab:robust}.
\end{tablenotes}
\end{threeparttable}
\end{table*}

\begin{table*}[pos=tbp]\centering\small
\begin{threeparttable}
\caption{Conformity during consultation.}\label{tab:mechanism}
\setlength{\tabcolsep}{3pt}
\begin{tabular}{@{}>{\raggedright\arraybackslash}p{24mm}lrrrrrr@{}}\dtoprule
\multicolumn{2}{@{}l}{} & \multicolumn{3}{c}{MedQA-USMLE} & \multicolumn{3}{c}{VM14K} \\
\cmidrule(lr){3-5}\cmidrule(l){6-8}
\multicolumn{2}{@{}l}{Measure} & Qwen3-32B & Llama-3.3-70B & Llama-3.1-8B & Qwen3-32B & Llama-3.3-70B & Llama-3.1-8B \\ \midrule
\multirow[t]{2}{=}{Adopts the majority answer} & If minority correct & 71.9 & 61.6 & 59.1 & 69.3 & 59.7 & 59.7 \\
 & If minority wrong & 76.4 & 68.4 & 77.7 & 76.2 & 66.6 & 63.0 \\
\midrule
\multicolumn{2}{@{}l}{Minority override} & 78.0 & 76.8 & 65.3 & 74.6 & 78.0 & 68.3 \\
 
\dbottomrule\end{tabular}
\begin{tablenotes}[flushleft]\footnotesize
\item[] VM14K, 2,000-question subset. Adopts the majority answer, share of cases in which the single dissenting agent of a two-to-one split adopts the majority answer, by whether its own answer was correct. Minority override, share of questions with exactly one correct pre-consultation vote on which the final answer is incorrect, as defined in Section~\ref{sec:metrics}.
\end{tablenotes}
\end{threeparttable}
\end{table*}

\begin{table*}[pos=tbp]\centering\small
\begin{threeparttable}
\caption{AUROC of post-consultation agreement under simulated revision rules.}\label{tab:null}
\setlength{\tabcolsep}{3pt}
\begin{tabular}{@{}>{\raggedright\arraybackslash}p{24mm}lrrrrrr@{}}\dtoprule
\multicolumn{2}{@{}l}{} & \multicolumn{3}{c}{MedQA-USMLE} & \multicolumn{3}{c}{VM14K} \\
\cmidrule(lr){3-5}\cmidrule(l){6-8}
\multicolumn{2}{@{}l}{Agreement} & Qwen3-32B & Llama-3.3-70B & Llama-3.1-8B & Qwen3-32B & Llama-3.3-70B & Llama-3.1-8B \\ \midrule
\multirow[t]{2}{=}{Observed} & Pre-consultation & 69.2 & 67.8 & 66.9 & 65.4 & 62.5 & 62.1 \\
 & Post-consultation & 54.3 & 57.3 & 57.2 & 53.6 & 55.0 & 56.1 \\
\midrule
\multirow[t]{3}{=}{Simulated post-consultation} & Observed switching rates & 55.4 & 58.0 & 57.4 & 55.2 & 55.3 & 55.0 \\
 & Indifferent to correctness & \tdci{55.0}{53.1}{57.1} & \tdci{56.9}{54.3}{59.5} & \tdci{54.8}{53.2}{56.7} & \tdci{54.4}{53.2}{55.6} & \tdci{54.8}{53.6}{55.9} & \tdci{54.7}{53.4}{56.3} \\
 & Only wrong minorities switch & \tdci{66.1}{66.0}{66.2} & \tdci{67.2}{67.0}{67.4} & \tdci{68.1}{68.1}{68.1} & \tdci{64.2}{64.2}{64.2} & \tdci{61.0}{60.9}{61.2} & \tdci{64.5}{63.8}{65.2} \\
\midrule
\multirow[t]{2}{=}{Unanimous after consultation (\%)} & Indifferent to correctness & 94.7 & 96.0 & 83.6 & 92.3 & 90.6 & 66.5 \\
 & Only wrong minorities switch & 94.2 & 95.8 & 82.7\tnote{a} & 91.1\tnote{a} & 90.6 & 66.6 \\
\dbottomrule\end{tabular}
\begin{tablenotes}[flushleft]\footnotesize
\item[] AUROC for an incorrect final answer, 0--100 scale; observed values from Table~\ref{tab:selective}. Only two-to-one questions are simulated, 200 runs per seed, and the final answer is the plurality of the post-consultation votes. Brackets give the 2.5th and 97.5th percentiles. Indifferent, one pooled switching rate for every minority agent. Only wrong switch, correct minority agents keep their answer.
\item[a] Switching rate capped at one.
\end{tablenotes}
\end{threeparttable}
\end{table*}

\subsection{Signals that remain informative}
\label{sec:res-league}

\begin{figure*}[pos=tbp]
\centering
\includegraphics[width=\linewidth]{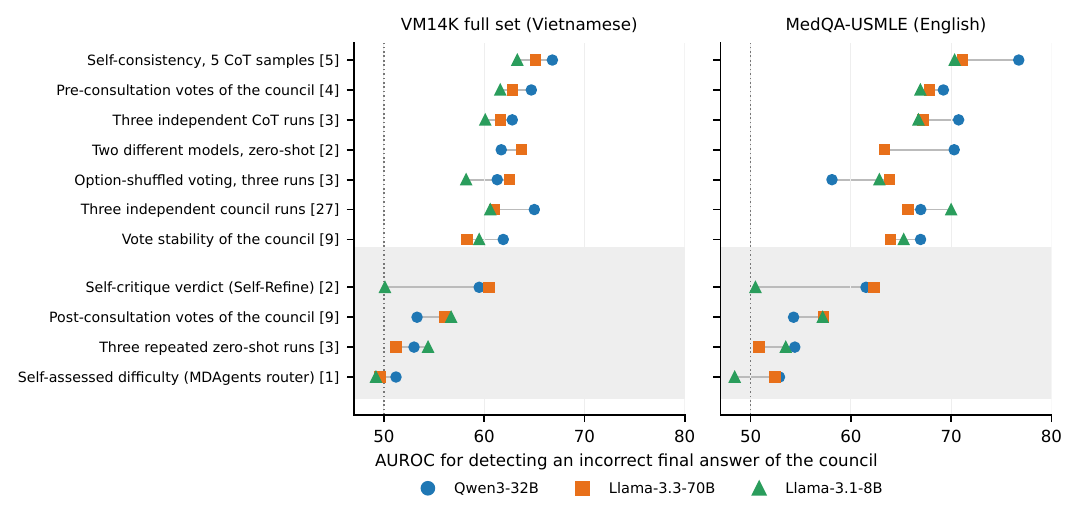}
\caption{Confidence signals of every strategy, ranked by AUROC for detecting an incorrect final answer of the protocol (50 is chance): signals from independent reasoning draws (top) against signals from repeated prompts, interaction or self-assessment (shaded); brackets give calls per question. Every signal is scored against the final answer of the protocol, not against the answer of its own strategy. VM14K: full set (Table~\ref{tab:S5}).}
\label{fig:league}
\end{figure*}

Every strategy we evaluated yields a confidence signal, and Fig.~\ref{fig:league} ranks these signals by their AUROC for detecting an incorrect final answer of the consultation protocol (Table~\ref{tab:S5}).
All signals are scored against this same answer, because the question is which signal a system that runs consultation should use to defer its own answer.
A signal from another strategy can therefore lose credit on questions on which that strategy and the protocol give different answers, so its AUROC here can be lower than for its own answers.
For five CoT-SC samples with Qwen3-32B on MedQA-USMLE, for example, it was 76.7 for the answer of the protocol and 78.1 for its own answers (Table~\ref{tab:selective}).
In every cell, post-consultation agreement ranked below every signal from independent reasoning draws that was evaluated in that cell.
These include three independent CoT runs and, for the two larger models, agreement between two different models.
One of these signals, agreement across three runs of the entire protocol, cost 27 calls and was not consistently better than the pre-consultation votes of a single run, which cost four.
Two signals were close to chance in every cell.
These were agreement among three repetitions of the same zero-shot prompt, with an AUROC of 50.9--54.4, and the question complexity that MDAgents assigns, with 48.4--52.9.

Two cheap signals stood out.
A run that records both sets of votes shows whether any agent changed its vote.
This binary signal reached an AUROC of 64.0--66.9 on MedQA-USMLE and 59.0--63.1 on VM14K, close to that of pre-consultation agreement.
Agreement between the zero-shot answers of Qwen3-32B and Llama-3.3-70B needs only two calls.
As a signal for the final answer of the protocol, this agreement reached an AUROC of 63.4--70.3 on MedQA-USMLE and 61.7--63.7 on the full VM14K set.

\subsection{Remedies and robustness}
\label{sec:res-remedies}\label{sec:res-robust}

Two changes keep a usable signal without any additional model.
The first records the votes before consultation.
Stopping after step~2 and answering with the plurality of the pre-consultation votes uses four of the nine calls.
On MedQA-USMLE, its agreement ranked its own answers with an AUROC of 67.6--67.7 (Table~\ref{tab:selective}), and it was unanimous and wrong on 5.4--11.2\% of questions, against 7.7--24.3\% for the full protocol.
An early exit that returns this answer when the three analyses agree, and completes the protocol only when they disagree, matched the accuracy of the full protocol to within 0.4 points in all six cells.
It needed 4.5--7.0 calls per question instead of nine, and pre-consultation agreement ranked its answers with an AUROC of 62--70, against 54--57 for post-consultation agreement (Table~\ref{tab:remedies}).

\begin{figure*}[pos=tbp]
\centering
\includegraphics[width=0.74\linewidth]{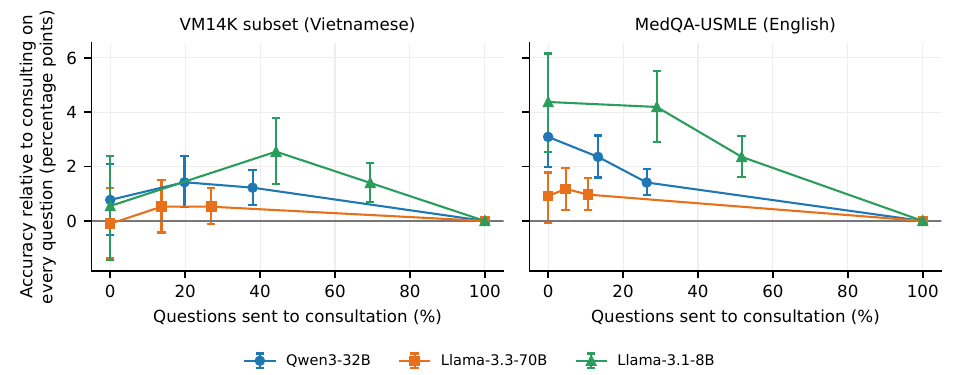}
\caption{Disagreement-gated consultation: accuracy relative to uniform consultation against the share of questions consulted, for gates requiring four of five and five of five CoT-SC samples to agree; 95\% paired bootstrap intervals. VM14K: 2,000-question subset.}
\label{fig:gating}
\end{figure*}

The second change keeps consultation but uses it only on disagreement.
CoT-SC answers when at least four of its five samples agree, and the remaining questions go to consultation (Table~\ref{tab:gating}, Fig.~\ref{fig:gating}).
On MedQA-USMLE this gate consulted on 5--29\% of questions, was 1.18--4.19 points more accurate than consulting on every question and did not differ from CoT-SC alone.
On VM14K it consulted on 14--44\% of questions and was 0.53--2.55 points more accurate than uniform consultation, although the interval for Llama-3.3-70B included zero.
It did not differ from CoT-SC except for Llama-3.1-8B, where it was 2.00 points more accurate, with a 95\% CI of 0.62 to 3.30.
Agreement among its CoT-SC samples ranked its answers with an AUROC of 65--80 (Table~\ref{tab:remedies}).

\begin{figure*}[pos=tbp]
\centering
\includegraphics[width=\linewidth]{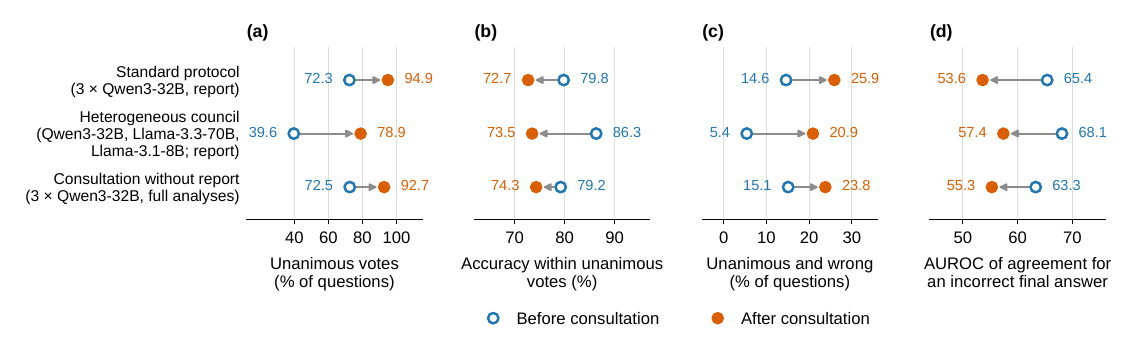}
\caption{Protocol variants on the VM14K subset: (a) unanimous votes, (b) accuracy of the unanimous answer, (c) unanimous and wrong and (d) AUROC of agreement for an incorrect final answer, before (open) and after (filled) consultation.}
\label{fig:variants}
\end{figure*}

The two protocol variants on VM14K reproduced the main findings (Table~\ref{tab:robust}, Fig.~\ref{fig:variants}).
Before consultation, the heterogeneous council was unanimous less often than three copies of Qwen3-32B, on 39.6\% of questions against 72.3\%, and its unanimous answers were more accurate, at 86.3\% against 79.8\%.
Consultation removed most of this difference.
Unanimity rose to 78.9\%, the accuracy of the unanimous answer fell to 73.5\%, and the AUROC of agreement fell from 68.1 to 57.4.
The council was 2.20 points less accurate than CoT-SC with Qwen3-32B alone, although at least one of its agents was correct on 85.7\% of questions.
The added agreement came mostly from the weakest member.
Over consultation, the accuracy of the votes of Llama-3.1-8B rose from 47.5\% to 63.1\%, while those of the two larger models changed by less than one point.
The final accuracy of the council, 69.7\%, was only 0.9 points above that of the single pre-consultation vote of Qwen3-32B.
When the agents read the three full analyses instead of the report, unanimity still rose from 72.5\% to 92.7\% and the AUROC still fell from 63.3 to 55.3, a change of $-$8.0 points.
Unanimous errors rose from 15.1\% to 23.8\%, 2.1 points fewer than with the report, and accuracy did not differ from CoT-SC at 2.8 times its tokens.
The loss therefore needs neither the report nor truncated inputs.

\begin{table*}[pos=tbp]\centering\small
\begin{threeparttable}
\caption{Early exit and disagreement-gated consultation as deferral systems.}\label{tab:remedies}
\setlength{\tabcolsep}{3pt}
\begin{tabular}{@{}>{\raggedright\arraybackslash}p{24mm}lrrrrrr@{}}\dtoprule
\multicolumn{2}{@{}l}{} & \multicolumn{3}{c}{MedQA-USMLE} & \multicolumn{3}{c}{VM14K} \\
\cmidrule(lr){3-5}\cmidrule(l){6-8}
\multicolumn{2}{@{}l}{System} & Qwen3-32B & Llama-3.3-70B & Llama-3.1-8B & Qwen3-32B & Llama-3.3-70B & Llama-3.1-8B \\ \midrule
\multirow[t]{4}{=}{Early exit on pre-consultation votes} & Accuracy (\%) & 83.3 & 90.7 & 67.3 & 71.0 & 69.4 & 49.5 \\
 & Calls & 4.9 & 4.5 & 6.2 & 5.4 & 5.0 & 7.0 \\
 & AURC ($\downarrow$) & 10.7 & 6.2 & 23.6 & 21.5 & 24.7 & 41.3 \\
 & AUROC ($\uparrow$) & 69.5 & 67.5 & 66.0 & 65.5 & 62.6 & 62.4 \\
\midrule
\multirow[t]{3}{=}{Gated consultation} & Accuracy (\%) & 85.6 & 91.9 & 71.3 & 72.5 & 69.9 & 51.7 \\
 & AURC ($\downarrow$) & 6.3 & 4.4 & 15.9 & 18.1 & 22.9 & 36.7 \\
 & AUROC ($\uparrow$) & 79.5 & 73.4 & 74.1 & 69.1 & 64.9 & 64.9 \\
 
\dbottomrule\end{tabular}
\begin{tablenotes}[flushleft]\footnotesize
\item[] VM14K, 2,000-question subset. Early exit returns the plurality of the pre-consultation votes when the three analyses agree and completes the protocol otherwise, and its answers are ranked by pre-consultation agreement. Gated consultation answers with CoT-SC when at least four of its five samples agree and consults otherwise, and its answers are ranked by CoT-SC agreement. Calls, mean model calls per question.
\end{tablenotes}
\end{threeparttable}
\end{table*}

\begin{table*}[pos=tbp]\centering\small
\begin{threeparttable}
\caption{Disagreement-gated consultation.}\label{tab:gating}
\setlength{\tabcolsep}{3pt}
\begin{tabular}{@{}>{\raggedright\arraybackslash}p{24mm}lrrrrrr@{}}\dtoprule
\multicolumn{2}{@{}l}{} & \multicolumn{3}{c}{MedQA-USMLE} & \multicolumn{3}{c}{VM14K} \\
\cmidrule(lr){3-5}\cmidrule(l){6-8}
\multicolumn{2}{@{}l}{Measure} & Qwen3-32B & Llama-3.3-70B & Llama-3.1-8B & Qwen3-32B & Llama-3.3-70B & Llama-3.1-8B \\ \midrule
\multirow[t]{3}{=}{Accuracy (\%)} & CoT-SC & \textbf{86.3} & \underline{91.7} & \textbf{71.5} & \underline{71.9} & 69.3 & \underline{49.7} \\
 & Consultation & 83.3 & 90.7 & 67.1 & 71.1 & \underline{69.4} & 49.1 \\
 & Gated consultation & \underline{85.6} & \textbf{91.9} & \underline{71.3} & \textbf{72.5} & \textbf{69.9} & \textbf{51.7} \\
\midrule
\multicolumn{2}{@{}l}{Consulted (\%)} & 13 & 5 & 29 & 20 & 14 & 44 \\
\midrule
\multirow[t]{2}{=}{Gated, difference (pp)} & vs consultation & \tdci{$+$2.36}{$+$1.60}{$+$3.14} & \tdci{$+$1.18}{$+$0.42}{$+$1.94} & \tdci{$+$4.19}{$+$2.91}{$+$5.50} & \tdci{$+$1.43}{$+$0.50}{$+$2.38} & \tdci{$+$0.53}{$-$0.43}{$+$1.50} & \tdci{$+$2.55}{$+$1.35}{$+$3.77} \\
 & vs CoT-SC & \tdci{$-$0.73}{$-$1.47}{$+$0.03} & \tdci{$+$0.26}{$-$0.24}{$+$0.79} & \tdci{$-$0.18}{$-$1.36}{$+$1.00} & \tdci{$+$0.65}{$-$0.15}{$+$1.50} & \tdci{$+$0.62}{$-$0.18}{$+$1.40} & \tdci{$+$2.00}{$+$0.62}{$+$3.30} \\
 
\dbottomrule\end{tabular}
\begin{tablenotes}[flushleft]\footnotesize
\item[] VM14K, 2,000-question subset. Gated consultation answers with CoT-SC when at least four of its five samples agree and consults otherwise. Bold, most accurate strategy in each column. Underlined, second most accurate. Differences are paired by question, with 95\% bootstrap CI in brackets.
\end{tablenotes}
\end{threeparttable}
\end{table*}

\begin{table*}[pos=tbp]\centering\small
\begin{threeparttable}
\caption{Protocol variants on VM14K.}\label{tab:robust}
\setlength{\tabcolsep}{6pt}
\begin{tabular}{@{}>{\raggedright\arraybackslash}p{34mm}lrrr@{}}\dtoprule
\multicolumn{2}{@{}l}{Measure} & \makecell{Standard\\protocol} & \makecell{Heterogeneous\\council} & \makecell{Consultation\\without report} \\ \midrule
\multirow[t]{2}{=}{Overall} & Accuracy (\%) & \tdci{71.1}{69.2}{72.8} & \tdci{69.7}{67.9}{71.5} & \tdci{72.0}{70.2}{73.8} \\
 & vs CoT-SC (pp) & \tdci{$-$0.78}{$-$2.10}{$+$0.50} & \tdci{$-$2.20}{$-$3.67}{$-$0.78} & \tdci{$+$0.10}{$-$1.35}{$+$1.45} \\
\midrule
\multirow[t]{2}{=}{Unanimous votes (\%)} & Before & 72.3 & 39.6 & 72.5 \\
 & After & 94.9 & 78.9 & 92.7 \\
\midrule
\multirow[t]{2}{=}{Accuracy of unanimous answer (\%)} & Before & 79.8 & 86.3 & 79.2 \\
 & After & 72.7 & 73.5 & 74.3 \\
\midrule
\multirow[t]{2}{=}{Unanimous and wrong (\%)} & Before & 14.6 & 5.4 & 15.1 \\
 & After & 25.9 & 20.9 & 23.8 \\
\midrule
\multirow[t]{3}{=}{AUROC of agreement} & Before & 65.4 & 68.1 & 63.3 \\
 & After & 53.6 & 57.4 & 55.3 \\
 & Change & \tdci{$-$11.8}{$-$14.1}{$-$9.4} & \tdci{$-$10.7}{$-$13.3}{$-$8.1} & \tdci{$-$8.0}{$-$10.8}{$-$5.2} \\
\midrule
\multirow[t]{7}{=}{Two-to-one questions} & Questions per seed & 500 & 988 & 501 \\
 & Initial majority (\%) & 48.1 & 62.7 & 52.7 \\
 & Final answer (\%) & \underline{49.7} & \underline{63.0} & \underline{54.3} \\
 & CoT-SC (\%) & \textbf{55.1} & \textbf{66.4} & \textbf{56.0} \\
 & Oracle (\%) & 79.0 & 84.5 & 81.2 \\
 & Final $-$ majority (pp) & \tdci{$+$1.6}{$-$0.6}{$+$3.6} & \tdci{$+$0.3}{$-$0.9}{$+$1.6} & \tdci{$+$1.6}{$-$0.2}{$+$3.4} \\
 & Final $-$ CoT-SC (pp) & \tdci{$-$5.5}{$-$8.9}{$-$2.3} & \tdci{$-$3.4}{$-$5.6}{$-$1.3} & \tdci{$-$1.7}{$-$5.3}{$+$1.5} \\
 
\dbottomrule\end{tabular}
\begin{tablenotes}[flushleft]\footnotesize
\item[] VM14K, 2,000-question subset, with Qwen3-32B as the standard model and two seeds. 95\% bootstrap CI in brackets, and differences are paired by question. AUROC of agreement for detecting an incorrect final answer. Two-to-one questions, questions split two to one before consultation, with the quantities of Table~\ref{tab:split}. Bold, most accurate of the initial majority, the final answer and CoT-SC in each column. Underlined, second most accurate. Mean tokens per question were 4,689 for the standard protocol, 3,872 for the heterogeneous council, 7,369 for consultation without report and 2,606 for CoT-SC with Qwen3-32B.
\end{tablenotes}
\end{threeparttable}
\end{table*}

The runs on all 11,094 questions of the full VM14K set, under the legacy configuration, reproduced every effect (\ref{app:config}, Table~\ref{tab:repl}).
For Qwen3-32B and Llama-3.3-70B, unanimity rose from 67.6\% and 77.7\% to 94.2\% and 90.3\%, unanimous errors rose from 14.8\% and 18.2\% to 27.0\% and 25.6\% of questions, and the AUROC of agreement fell from 64.7 and 62.8 to 53.3 and 56.1.
Unanimity and unanimous errors rose in each of the 30 specialties with at least 150 questions, for both models (Fig.~\ref{fig:A2}).
For Qwen3-32B the effects were larger on the 1,223 questions labelled challenging or hard, where unanimity rose from 60.7\% to 94.4\% and the AUROC fell from 63.0 to 52.9 (Table~\ref{tab:sens}b).
Excluding 192 questions that are candidates for label errors, on which at least 90\% of the 27 answers from the three models and three strategies chose the same option other than the reference answer, did not change the direction of any result (Table~\ref{tab:sens}c).

\section{Discussion}
\label{sec:discussion}

\subsection{Agreement after interaction is not a measure of confidence}
\label{sec:disc-agreement}

Agreement among answers tells us something about correctness only to the extent that the answers were produced independently.
The pre-consultation votes of the protocol meet this condition approximately, and their agreement was informative.
Its AUROC was similar to that of three independent CoT runs drawn at the same temperature and scored against the same answer, 66.9--69.2 against 66.7--70.7 on MedQA-USMLE (Table~\ref{tab:S5}), so role conditioning did not clearly weaken the signal before the agents interacted.
Self-consistency ranked higher, but its samples were drawn at a higher temperature, so this study cannot say how much of its margin comes from the number of draws and how much from temperature.
The post-consultation votes do not meet the condition.
Each was cast after reading a report built from all three analyses, and the votes moved towards the majority of those analyses whether or not the majority was correct (Table~\ref{tab:mechanism}).
The result is a signal that is high almost everywhere and says little about which answers are wrong, while the accuracy of the system does not improve.
We observed the same pattern for three models between 8 and 70 billion parameters, in English and in Vietnamese, and across the medical specialties examined.

The comparison across strategies places this result in a broader pattern.
Signals that rest on independent draws of the reasoning were informative, whether the draws were samples from one model, runs of the whole protocol or answers from a second model.
Agreement between two models still has a limit, because errors are correlated across models \cite{kim2025correlated}, so it says little on questions that both models answer wrongly.
Signals produced after the agents had exchanged information, by repeating a nearly deterministic prompt, or by asking the model to rate the difficulty of the question were weak or uninformative.
This agrees with the argument that consensus-seeking interaction filters out informative disagreement \cite{chen2025colmad}, with the reduced discrimination of self-reported confidence under debate \cite{bertalanic2026cost}, and with the chance-level terminal agreement reported for a multi-round medical consensus system, whose first-round agreement discriminated at a level close to the pre-consultation agreement measured here \cite{wang2026unanimously}.
In our setting the loss needs no stopping rule and occurs within a single step of interaction.
In clinical terms, a system that defers on post-consultation agreement sends one question in five to a clinician while barely enriching the deferred questions for errors.

\subsection{Relation to conformity and to accuracy}
\label{sec:disc-conformity}

That agents conform is well established \cite{wynn2025talk,yao2025peacemaker,bertalanic2026cost,gu2025medagentaudit}.
Our data add two observations.
A minority agent whose answer was correct still adopted the majority answer in more than half of the cases, at a rate only 5 to 19 points below that of a minority agent whose answer was wrong on MedQA-USMLE.
The votes moved mainly towards the majority and only secondarily towards the correct answer.
The simulation in Section~\ref{sec:res-mechanism} shows that this indifference, not convergence itself, is enough to explain the loss.
Social influence has the same effect in human groups, narrowing estimates and raising confidence without improving accuracy \cite{lorenz2011social}.
The cost of this conformity shows less in accuracy, which changed by less than one point relative to the initial majority on MedQA-USMLE, than in the loss of the information that the disagreement carried.
The report summarises points of agreement and disagreement, so it tells each agent which answer most analyses chose.
This is one source of conformity that prompt design could reduce.
The loss also occurred when agents read the full analyses instead of the report.
Consultation also did not turn this information into accuracy where it would have helped most.
On the questions on which the analyses disagreed, the final answer stayed within a few points of the initial majority and below self-consistency, even though a correct analysis was present in most of them (Table~\ref{tab:split}).
An evaluation of a consultation system by its final accuracy alone therefore misses what consultation removes.

\subsection{Implications for the design and evaluation of consultation systems}
\label{sec:disc-implications}

Three practical consequences follow for systems of this kind.
First, the votes cast before any interaction should be recorded and exposed, and deferral should be based on them rather than on the agreement reported at the end of the consultation.
The Delphi method follows the same principle and collects expert judgements independently before any feedback \cite{dalkey1963experimental}.
Second, where consultation is retained, it is better used as a conditional step restricted to the questions on which independent samples disagree, whether those samples are the five draws of self-consistency or the three analyses of the protocol itself.
Gating limits the losses of uniform consultation but adds little over self-consistency.
On MedQA-USMLE the gated system was no more accurate than self-consistency alone, which needs no consultation.
Third, evaluations of multi-agent medical systems should report a selective-prediction metric for the confidence signal that the system exposes, alongside accuracy.
A unanimous council should not be presented to a clinician as a confident one without such evidence.

Neither of two natural modifications restores the signal.
A council of three different models has lower initial agreement and more reliable unanimity, but after consultation its unanimous answers were no more accurate than those of three copies of one model, 73.5\% against 72.7\%, and most of the added agreement came from the smallest model.
Giving the agents one another's full analyses instead of a summary reduces the inflation of agreement by a few points and leaves the loss of discrimination in place, at nearly three times the cost of self-consistency.
The three consequences above therefore hold for both variants.
The signal is masked rather than destroyed: it is present in the same run at no additional cost, and recording the votes cast before any exchange between agents is enough to recover it.

\section{Limitations and future work}
\label{sec:limitations}

This study has several limitations.
We examined one single-round protocol that passes on only the opening of each analysis, and its two variants were run only on the VM14K subset with Qwen3-32B.
In the heterogeneous council, the model that wrote the report and made the decision was also an agent.
The models were open-weight models of 8 to 70 billion parameters with 4-bit quantisation, and both datasets contain four-option examination questions that no clinician reviewed.
MedQA-USMLE may be partly memorised, although every effect recurred on VM14K, released in 2025.
CoT-SC was sampled at a higher temperature than the other calls, which does not affect the comparison of pre- with post-consultation votes but blurs the margin of self-consistency.
All prompts were in English, and the VM14K labels have not been independently audited.
The full VM14K set was run under a legacy configuration; it is the only set with every strategy and underlies Fig.~\ref{fig:league}.

Future work should test whether the loss of the signal holds for protocols with several rounds or other decision rules, since conformity has been reported to grow over rounds of debate \cite{wynn2025talk,yao2025peacemaker,bertalanic2026cost}.
It should also test prompts that ask agents to keep a minority answer unless they find a specific error \cite{pitre2025consensagent,wang2025silence}.
Proprietary and reasoning models, open-ended clinical cases judged by clinicians, and reports written in Vietnamese for Vietnamese questions are further directions.
Comparing CoT-SC and consultation at the same sampling temperature would also separate the effect of independent sampling from that of temperature.

\section{Conclusion}
\label{sec:conclusion}

In a consultation protocol for medical question answering, the agreement of the agents was an informative deferral signal before they consulted and a nearly uninformative one afterwards, for three models and two languages, while accuracy on MedQA-USMLE remained below that of five-sample self-consistency, which uses fewer calls.
Unanimity reached through consultation mixed questions that had been agreed from the start with questions on which a council split two to one had converged, and the latter were answered correctly in at most 63\% of cases.
The same loss occurred when the three agents were three different models and when they read one another's full analyses instead of a report.
On the questions on which the agents disagreed, consultation added little to a majority vote over the analyses it started from.
For systems of this kind, the votes cast before interaction are the ones that indicate when a question should be deferred to a clinician, and consultation, where it is retained, is best reserved for the questions on which independent samples disagree.

\section*{Declaration of competing interest}
The authors declare that they have no known competing financial interests or personal relationships that could have appeared to influence the work reported in this paper.

\section*{Ethics statement}
This study used only publicly released question sets (MedQA-USMLE and VM14K) and involved no patient data and no human participants.

\section*{Data and code availability}
MedQA-USMLE \cite{jin2021medqa} and VM14K \cite{nguyen2025vm14k} are publicly available from their original sources under their respective licenses.
The code, the prompts, the identifiers of the VM14K questions used, the outputs and vote logs of all runs, and the scripts that generate every table and figure in this paper are available at \url{https://github.com/phenikaa-a2i-lab/MACE}.

\section*{Declaration of generative AI and AI-assisted technologies in the manuscript preparation process}
During the preparation of this work the authors used Claude (Anthropic) to improve the language and readability of the manuscript.
After using this tool, the authors reviewed and edited the content as needed and take full responsibility for the content of the publication.

\section*{Funding}
This research did not receive any specific grant from funding agencies in the public, commercial, or not-for-profit sectors.

\printcredits

\appendix
\setcounter{table}{0}\setcounter{figure}{0}
\renewcommand{\thetable}{A.\arabic{table}}\renewcommand{\thefigure}{A.\arabic{figure}}
\renewcommand{\thesection}{Appendix~\Alph{section}}
\section{Experimental configurations, vote extraction and replication on the full VM14K set}
\label{app:config}

\paragraph{Data, seeds and resampling.}
The 11,094 four-option questions of VM14K comprise 3,556 easy, 6,315 medium, 1,133 challenging and 90 hard questions, and the 2,000-question subset comprises 650 easy, 1,131 medium, 207 challenging and 12 hard questions.
Runs with three seeds used seeds 42, 43 and 44, and runs on the VM14K subset used seeds 42 and 43.
Bootstrap intervals used 1,000 resamples for accuracy and 2,000 for paired differences in accuracy, and two-sided $p$-values used 10,000 resamples.
Intervals for differences in AURC and AUROC used 1,000 resamples on MedQA-USMLE and the VM14K subset and 300 on the full VM14K set.

\paragraph{Configurations.}
The primary runs on MedQA-USMLE and the VM14K subset used SGLang with 4-bit AWQ checkpoints, an output limit of 1,024 tokens per call and complete logging of raw outputs.
The runs on the full VM14K set used a legacy configuration with output limits of 512 tokens for CoT, 600 for an analysis and 300 for the report and each revision.
The serving stack (Ollama with GGUF Q4\_K\_M, LiteLLM with AWQ, or SGLang with AWQ) varied by model and strategy, and raw outputs were stored only for Llama-3.3-70B and for the topic-aware consultation run of Llama-3.1-8B.
The lower output limit had a measurable effect.
It truncated 13.0\% of the CoT outputs of Qwen3-32B and 5.7\% of those of Llama-3.3-70B, and the truncated outputs were correct in about 27\% of cases (Table~\ref{tab:limit}); on MedQA-USMLE, with a limit of 1,024 tokens, at most 1.7\% of CoT outputs reached it.
The serving stack had a small effect.
Where the same questions were answered on two stacks, accuracy differed by $+$0.19 points [$-$2.12; $+$2.54] for CoT with Qwen3-32B (1,062 questions), $-$1.28 [$-$2.11; $-$0.44] for CoT-SC with Llama-3.1-8B (10,827 questions) and $-$0.40 [$-$1.35; $+$0.61] for the topic-aware consultation of Llama-3.1-8B (11,094 questions), and the answers given on two stacks agreed 2--6 points less often than the answers of two seeds on the same stack.

\begin{table*}[pos=tbp]\centering\small
\begin{threeparttable}
\caption{Effect of the CoT output limit.}\label{tab:limit}
\setlength{\tabcolsep}{4pt}
\begin{tabular}{@{}>{\raggedright\arraybackslash}p{24mm}lrrrrr@{}}\dtoprule
\multicolumn{2}{@{}l}{} & \multicolumn{3}{c}{MedQA-USMLE} & \multicolumn{2}{c}{VM14K full set} \\
\cmidrule(lr){3-5}\cmidrule(l){6-7}
\multicolumn{2}{@{}l}{Measure} & Qwen3-32B & Llama-3.3-70B & Llama-3.1-8B & Qwen3-32B & Llama-3.3-70B \\ \midrule
\multicolumn{2}{@{}l}{Outputs} & 3,816 & 3,816 & 3,816 & 1,062\tnote{a} & 33,282 \\
\multicolumn{2}{@{}l}{At limit (\%)} & 1.7 & 0.1 & 1.6 & 13.0 & 5.7 \\
\midrule
\multirow[t]{3}{=}{Accuracy (\%)} & At limit & 12.1 & 0.0 & 0.0 & 26.8 & 27.4 \\
 & Below limit & 84.2 & 90.9 & 67.9 & 71.9 & 69.7 \\
 & Overall & 82.9 & 90.9 & 66.8 & 66.0 & 67.3 \\
\dbottomrule\end{tabular}
\begin{tablenotes}[flushleft]\footnotesize
\item[] The limit was 1,024 tokens on MedQA-USMLE and 512 tokens on the full VM14K set. Outputs, CoT outputs with logged token counts. At limit, share of these outputs that reached the limit. Accuracy of the outputs at the limit, below it and overall. Llama-3.1-8B is not shown for the full VM14K set because no token counts were logged.
\item[a] Token counts were logged for only part of the outputs.
\end{tablenotes}
\end{threeparttable}
\end{table*}

\paragraph{Vote extraction and re-executed runs.}
The option letter of each analysis and revision was extracted at run time by a rule-based parser.
Because its coverage proved incomplete, we re-extracted all votes from the stored raw outputs with a single rule: the last explicit, non-negated statement of an answer in the output, such as ``the answer is~X'', ``\textit{đáp án là}~X'' or an option letter followed by the option text.
Table~\ref{tab:votes} lists, for every run, the share of turns without a run-time vote, the share still without a vote after re-extraction, and the share of votes to which the two procedures assigned different letters.
Where an incorrectly parsed vote had been passed to a later step, that step was re-executed from the stored analyses with the corrected votes.
This concerned the decision step of the heterogeneous council and steps~3--5 for the two Llama models on MedQA-USMLE.
The re-execution changed the final answer on 23 of 4,000 question--seed pairs in the heterogeneous council and on 1.9--2.5\% (Llama-3.3-70B) and 12.6--15.1\% (Llama-3.1-8B) of pairs on MedQA-USMLE, and it changed mean accuracy by at most 0.6 percentage points.
All reported results of these runs are those of the re-executed protocol; token counts refer to the original nine-call runs.
In every run for which the finish reason of each call was logged, at most 0.25\% of the consultation calls reached the output limit, so Table~\ref{tab:votes} reflects the parser rather than truncation.
On the full VM14K set, raw outputs of Qwen3-32B were not stored; its run-time votes are used and marked \textsuperscript{b} wherever they appear.

\begin{table*}[pos=tbp]\centering\small
\begin{threeparttable}
\caption{Vote extraction.}\label{tab:votes}
\setlength{\tabcolsep}{4pt}
\begin{tabular}{@{}>{\raggedright\arraybackslash}p{25mm}>{\raggedright\arraybackslash}p{50mm}rrrrrr@{}}\dtoprule
\multicolumn{2}{@{}l}{} & \multicolumn{2}{c}{Analysis without vote} & \multicolumn{2}{c}{Revision without vote} & \multicolumn{2}{c}{Different letter} \\
\cmidrule(lr){3-4}\cmidrule(lr){5-6}\cmidrule(l){7-8}
\multicolumn{2}{@{}l}{Run} & Run time & Re-extracted & Run time & Re-extracted & Analysis & Revision \\ \midrule
\multirow[t]{3}{=}{MedQA-USMLE} & Qwen3-32B & 1.3--1.6 & 0.1--0.2 & 0.1--0.5 & 0.1 & 0.0--0.3 & 1.7--2.8 \\
 & Llama-3.3-70B & 1.6--1.8 & 0.1--0.2 & 0.4--0.6 & 0.1 & 0.3--0.4 & 1.0--1.4 \\
 & Llama-3.1-8B & 10.5--11.7 & 0.0--0.2 & 3.0--3.7 & 1.5--1.9 & 3.4--4.3 & 1.5--2.0 \\
\midrule
\multirow[t]{5}{=}{VM14K} & Qwen3-32B & 7.0--7.2 & 0.0--0.1 & 0.3--0.4 & 0.1 & 0.2 & 1.9--2.2 \\
 & Llama-3.3-70B & 3.3--4.0 & 0.2--0.3 & 1.3 & 0.2--0.3 & 2.6--2.7 & 2.4 \\
 & Llama-3.1-8B & 6.5--6.7 & 0.3--0.4 & 8.4--8.7 & 1.7--1.8 & 5.4--5.5 & 6.9 \\
 & Heterogeneous council & 1.8 & 0.1 & 16.8--17.9 & 0.5--0.6 & 2.6--3.1 & 2.6 \\
 & Consultation without report & 7.1--7.4 & 0.0--0.1 & 0.7--0.9 & 0.1 & 0.2--0.3 & 1.6--2.0 \\
\midrule
\multirow[t]{3}{=}{VM14K full set} & Llama-3.3-70B & 0.0 & 0.0 & 0.1 & 0.1 & 10.2--10.3 & 9.3--9.8 \\
 & Llama-3.1-8B, topic-aware\tnote{b} & 0.2 & 0.2 & 0.7 & 0.7 & 13.6 & 12.6 \\
 & Qwen3-32B\tnote{a} & 0.1 & -- & 0.0 & -- & -- & -- \\
\dbottomrule\end{tabular}
\begin{tablenotes}[flushleft]\footnotesize
\item[] Percentages of turns, as ranges over seeds. Without vote, share of turns without a vote at run time and after re-extraction. Different letter, share of votes to which the two procedures assigned different letters. MedQA-USMLE and VM14K use the unified configuration with three and two seeds, and the full VM14K set uses the legacy configuration. Under the unified configuration the run-time vote is kept when present and the re-extracted vote is used otherwise. Under the legacy configuration the re-extracted vote is used wherever raw outputs were stored.
\item[a] Raw outputs not stored.
\item[b] Consultation with topic-aware roles, seed 44.
\end{tablenotes}
\end{threeparttable}
\end{table*}

\paragraph{Replication on the full VM14K set.}
Table~\ref{tab:repl} repeats the main analyses on the 11,094 questions for Qwen3-32B and Llama-3.3-70B.
The accuracies of all strategies are given in Table~\ref{tab:S3}, the comparison of confidence signals in Table~\ref{tab:S5}, and the sensitivity to difficulty and label noise in Table~\ref{tab:sens}.
Every effect reported in the main text recurs.
Consultation raised unanimity and unanimous errors and lowered the AUROC of agreement to 53--56.
On the questions split two to one it added 1.8 and 4.6 points to the initial majority without improving on CoT-SC; the larger gain for Qwen3-32B reflects its run-time votes, in which a missing vote counts as dissent and the initial majority is therefore underestimated.
The gated policy was more accurate than uniform consultation for all three models.
The agreement statistics of Llama-3.3-70B match those of the subset to within two points on every quantity, and those of Qwen3-32B to within five points before consultation and two points after it; the remaining gap for Qwen3-32B reflects the run-time parser and the 600-token analysis limit of the legacy configuration.
The one quantity that changes sign is the accuracy of consultation relative to CoT-SC for Qwen3-32B ($+$2.00 points [$+$1.42; $+$2.65]).
This difference is attributable to the 512-token limit of the legacy CoT baseline, under which 13.0\% of the CoT outputs of this model were truncated and correct in 26.8\% of cases against 71.9\% for the others, and it disappears on the subset.

\begin{table*}[pos=tbp]\centering\small
\begin{threeparttable}
\caption{Replication on the full VM14K set.}\label{tab:repl}
\setlength{\tabcolsep}{4pt}
\begin{tabular}{@{}>{\raggedright\arraybackslash}p{34mm}>{\raggedright\arraybackslash}p{32mm}rr@{}}\dtoprule
\multicolumn{2}{@{}l}{Measure} & Qwen3-32B\tnote{a} & Llama-3.3-70B \\ \midrule
\multirow[t]{2}{=}{Unanimous votes (\%)} & Before & 67.6 & 77.7 \\
 & After & 94.2 & 90.3 \\
\midrule
\multirow[t]{2}{=}{Accuracy of unanimous answer (\%)} & Before & 78.1 & 76.5 \\
 & After & 71.4 & 71.7 \\
\midrule
\multirow[t]{2}{=}{Unanimous and wrong (\%)} & Before & 14.8 & 18.2 \\
 & After & 27.0 & 25.6 \\
\midrule
\multirow[t]{3}{=}{Accuracy (\%)} & Initial majority & 67.6 & 69.3 \\
 & Final answer & 69.9 & 68.9 \\
 & Oracle & 78.4 & 76.8 \\
\midrule
\multirow[t]{2}{=}{AURC} & Pre-consultation & 22.7 & 24.8 \\
 & Post-consultation & 28.8 & 28.3 \\
\midrule
\multirow[t]{3}{=}{AUROC} & Pre-consultation & 64.7 & 62.8 \\
 & Post-consultation & 53.3 & 56.1 \\
 & Change & \tdci{$-$11.5}{$-$12.6}{$-$10.4} & \tdci{$-$6.8}{$-$7.7}{$-$5.7} \\
\midrule
\multirow[t]{2}{=}{CoT-SC} & AURC & 19.7 & 22.3 \\
 & AUROC & 72.3 & 67.4 \\
\midrule
\multicolumn{2}{@{}l}{Error without deferral (\%)} & 30.1 & 31.1 \\
\midrule
\multicolumn{2}{@{}l}{Not unanimous before (\%)} & 28.8 & 16.4 \\
\midrule
\multirow[t]{3}{=}{Final accuracy (\%) by split before consultation} & 3--0 & 78.9 & 76.5 \\
 & 2--1 & 53.7 & 48.3 \\
 & 1--1--1 & 43.1 & 36.5 \\
\midrule
\multirow[t]{2}{=}{Answer changes on re-run (\%)} & Pre-consultation votes & 4.6 & 2.6 \\
 & Council & 12.4 & 7.5 \\
\midrule
\multirow[t]{2}{=}{Adopts the majority answer (\%)} & If minority correct & 62.1 & 54.9 \\
 & If minority wrong & 77.6 & 64.1 \\
\midrule
\multicolumn{2}{@{}l}{Minority override (\%)} & 68.6 & 73.9 \\
\midrule
\multirow[t]{6}{=}{Two-to-one questions} & Questions per seed & 3,193 & 2,579 \\
 & Initial majority (\%) & 47.9 & 50.5 \\
 & Final answer (\%) & 52.4 & 52.3 \\
 & CoT-SC (\%) & 52.5 & 53.7 \\
 & Final $-$ majority (pp) & \tdci{$+$4.6}{$+$3.8}{$+$5.3} & \tdci{$+$1.8}{$+$1.0}{$+$2.7} \\
 & Final $-$ CoT-SC (pp) & \tdci{$-$0.1}{$-$1.2}{$+$1.1} & \tdci{$-$1.4}{$-$2.5}{$-$0.3} \\
\midrule
\multicolumn{2}{@{}l}{Consulted (\%)} & 28 & 16 \\
\midrule
\multirow[t]{2}{=}{Gated, difference (pp)} & vs consultation & \tdci{$+$0.78}{$+$0.39}{$+$1.18} & \tdci{$+$0.80}{$+$0.40}{$+$1.21} \\
 & vs CoT-SC & \tdci{$+$2.79}{$+$2.38}{$+$3.20} & \tdci{$+$0.67}{$+$0.38}{$+$0.97} \\
 
\dbottomrule\end{tabular}
\begin{tablenotes}[flushleft]\footnotesize
\item[] 11,094 questions, three seeds, legacy configuration. The blocks follow Tables~\ref{tab:agreement}, \ref{tab:selective}, \ref{tab:strata}, \ref{tab:mechanism}, \ref{tab:split} and~\ref{tab:gating}. Pre- and post-consultation agreement rank the final answer of the protocol. Gated consultation answers with CoT-SC when at least four of its five samples agree. Differences are paired by question, with 95\% bootstrap CI in brackets. Llama-3.1-8B is not shown because its raw outputs on the full set exist only for the topic-aware variant; its gated policy consulted on 49\% of questions and differed from uniform consultation by $+$2.75 [$+$2.34, $+$3.22] and from CoT-SC by $-$0.22 [$-$0.74, $+$0.29] points.
\item[a] Run-time votes.
\end{tablenotes}
\end{threeparttable}
\end{table*}

\section{Prompts of the consultation protocol}
\label{app:prompts}

Table~\ref{tab:prompts} gives the prompts of the five steps of the protocol.
The fields \texttt{\{question\}}, \texttt{\{options\}}, \texttt{\{role\}} and \texttt{\{report\}} are filled at run time, and the bracketed items describe the content inserted by the implementation.

\begin{table*}[pos=tbp]\centering\small
\caption{Prompts of the consultation protocol, used in all runs. Step~1 has no system prompt.}\label{tab:prompts}
\renewcommand{\arraystretch}{1.15}
\begin{tabular}{@{}>{\raggedright\arraybackslash}p{0.11\linewidth}>{\raggedright\arraybackslash}p{0.24\linewidth}>{\raggedright\arraybackslash}p{0.60\linewidth}@{}}\dtoprule
Step & System prompt & User prompt \\ \midrule
1. Role selection & -- & You are coordinating a medical consultation. Medical question: \{question\} Options: \{options\} Which medical specialists would be most relevant to answer this question? List exactly 3 specialist roles, comma-separated. Example: Cardiologist, Nephrologist, General Practitioner. Only output the comma-separated list, nothing else. \\[2pt]
2. Analysis & You are an experienced \{role\}. Analyze medical questions carefully using your specialist expertise. & Medical question: \{question\} Options: \{options\} As a \{role\}, provide: 1. Your analysis of the question 2. Your evaluation of each option (A, B, C, D) 3. Your answer (A/B/C/D). Be concise and systematic. \\[2pt]
3. Report & You are a medical coordinator summarizing specialist consultations. & Medical question: \{question\} Options: \{options\} Specialist analyses: [for each role: role, extracted answer, opening of the analysis up to 500 characters] Summarize the key points of agreement and disagreement into a concise consultation report (3-5 sentences). \\[2pt]
4. Consultation & You are an experienced \{role\} participating in a multi-specialist consultation. & Medical question: \{question\} Options: \{options\} Consultation report from all specialists: \{report\} Your previous analysis: [opening of the agent's own analysis up to 500 characters] Given the collective consultation, do you maintain or update your answer? State your final answer (A/B/C/D) and brief reasoning. \\[2pt]
5. Decision & You are a senior physician making the final clinical decision. & Medical question: \{question\} Options: \{options\} Final specialist positions: [for each role: role, extracted answer, opening of the revision up to 500 characters] Based on the collective consultation, what is the final answer? Respond with only 1 character: A, B, C, or D. \\
\dbottomrule\end{tabular}
\end{table*}

\section{Additional results}
\label{app:results}
\begin{table*}[pos=tbp]\centering\small
\begin{threeparttable}
\caption{Accuracy of all strategies.}\label{tab:S3}
\setlength{\tabcolsep}{3pt}
\begin{tabular}{@{}>{\raggedright\arraybackslash}p{42mm}rrrrrr@{}}\dtoprule
 & \multicolumn{3}{c}{MedQA-USMLE} & \multicolumn{3}{c}{VM14K full set} \\
\cmidrule(lr){2-4}\cmidrule(l){5-7}
Strategy & Qwen3-32B & Llama-3.3-70B & Llama-3.1-8B & Qwen3-32B & Llama-3.3-70B & Llama-3.1-8B \\ \midrule
Zero-shot & \tdci{73.8}{71.3}{76.2} & \tdci{86.8}{85.0}{88.6} & \tdci{58.3}{55.9}{60.9} & \tdci{68.0}{67.1}{68.7} & \tdci{65.9}{65.0}{66.7} & \tdci{43.6}{42.8}{44.5} \\
CoT & \tdci{82.9}{81.2}{84.7} & \tdci{\underline{90.9}}{89.4}{92.2} & \tdci{66.8}{64.6}{68.9} & \tdci{64.8}{64.0}{65.5} & \tdci{67.3}{66.5}{68.1} & \tdci{\underline{47.8}}{47.1}{48.5} \\
CoT-SC & \tdci{\textbf{86.3}}{84.6}{88.1} & \tdci{\textbf{91.7}}{90.3}{93.0} & \tdci{\textbf{71.5}}{69.3}{73.6} & \tdci{67.9}{67.1}{68.7} & \tdci{\textbf{69.0}}{68.1}{69.8} & \tdci{\textbf{50.7}}{49.9}{51.5} \\
Self-Refine & \tdci{79.2}{77.3}{81.0} & \tdci{87.8}{86.5}{89.3} & \tdci{56.1}{53.9}{58.1} & \tdci{63.7}{63.0}{64.4} & \tdci{67.0}{66.3}{67.7} & \tdci{42.1}{41.4}{42.7} \\
Option-shuffled voting & \tdci{75.0}{73.0}{77.1} & \tdci{86.5}{84.7}{88.1} & \tdci{58.9}{56.6}{61.1} & \tdci{66.7}{66.0}{67.6} & \tdci{65.4}{64.7}{66.3} & \tdci{43.6}{42.9}{44.4} \\
Consultation & \tdci{\underline{83.3}}{81.3}{85.2} & \tdci{90.7}{89.3}{92.2} & \tdci{\underline{67.1}}{65.0}{69.2} & \tdci{\underline{69.9}}{69.1}{70.7} & \tdci{\underline{68.9}}{68.0}{69.7} & \tdci{47.7}{46.9}{48.5} \\
Consultation, topic-aware roles & -- & -- & -- & \tdci{\textbf{70.1}}{69.3}{70.8} & \tdci{68.7}{67.9}{69.5} & \tdci{47.5}{46.8}{48.2} \\
 
\dbottomrule\end{tabular}
\begin{tablenotes}[flushleft]\footnotesize
\item[] Accuracy (\%), mean of three seeds with the 95\% bootstrap CI in brackets below each value. The full VM14K set was run under the legacy configuration (\ref{app:config}). --, not run. Bold, most accurate strategy in each column. Underlined, second most accurate. Self-Refine, option-shuffled voting and the topic-aware variant were run for the confidence signals they give (Section~\ref{sec:res-league}), not as accuracy baselines. MDAgents was used only for its complexity estimate, so its accuracy is not reported.
\end{tablenotes}
\end{threeparttable}
\end{table*}

\begin{figure*}[pos=tbp]
\centering
\includegraphics[width=0.78\linewidth]{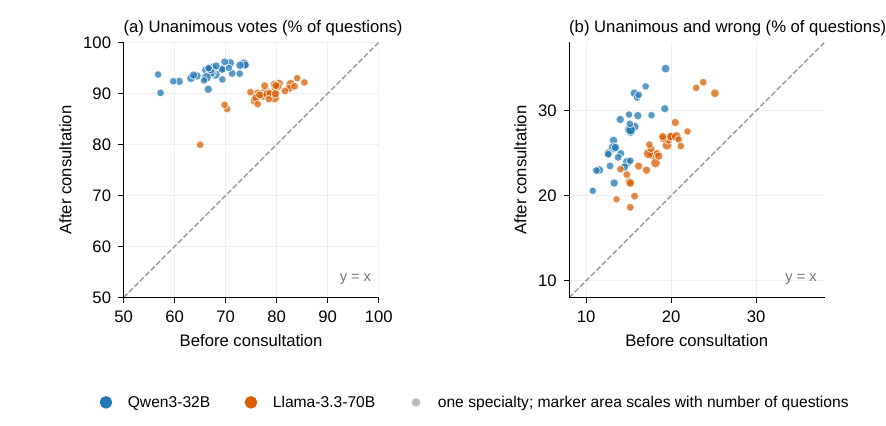}
\caption{Agreement by specialty on the full VM14K set. Each point is one of the 30 specialties with at least 150 questions (a question counts under each of its labels; marker area scales with the number of questions), plotted by its value after consultation against its value before, for (a) the share of questions with a unanimous vote and (b) the share of questions unanimous and wrong. Every point lies above the diagonal for both models.}
\label{fig:A2}
\end{figure*}

\begin{figure}[pos=tbp]
\centering
\includegraphics[width=\linewidth]{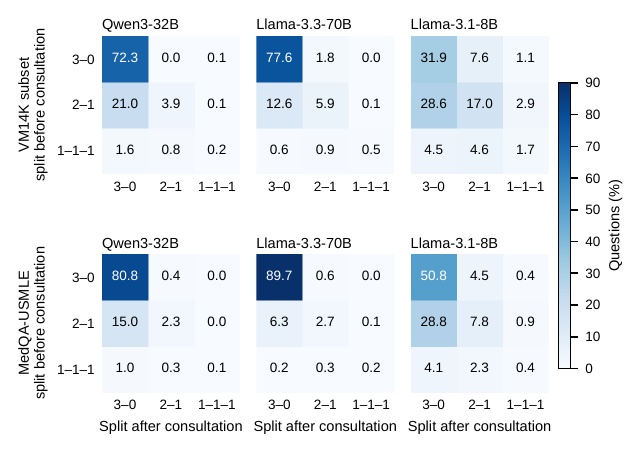}
\caption{Vote-split transitions: percentage of questions by split before (rows) and after (columns) consultation; unextracted votes count as dissenting.}
\label{fig:A1}
\end{figure}

Table~\ref{tab:S3} gives the accuracy of every strategy evaluated, Table~\ref{tab:S5} the values plotted in Fig.~\ref{fig:league}, Table~\ref{tab:sens} the three sensitivity analyses of Section~\ref{sec:res-robust}, Fig.~\ref{fig:A1} the vote-split transitions and Fig.~\ref{fig:A2} the specialty analysis.

\begin{table*}[pos=tbp]\centering\small
\begin{threeparttable}
\caption{AUROC of every confidence signal for detecting an incorrect final answer of the protocol.}\label{tab:S5}
\setlength{\tabcolsep}{3pt}
\begin{tabular}{@{}>{\raggedright\arraybackslash}p{38mm}rrrrrrrrrr@{}}\dtoprule
 & & \multicolumn{3}{c}{MedQA-USMLE} & \multicolumn{3}{c}{VM14K} & \multicolumn{3}{c}{VM14K full set} \\
\cmidrule(lr){3-5}\cmidrule(lr){6-8}\cmidrule(l){9-11}
Signal & Calls & \makecell{Qwen3\\32B} & \makecell{Llama-3.3\\70B} & \makecell{Llama-3.1\\8B} & \makecell{Qwen3\\32B} & \makecell{Llama-3.3\\70B} & \makecell{Llama-3.1\\8B} & \makecell{Qwen3\\32B} & \makecell{Llama-3.3\\70B} & \makecell{Llama-3.1\\8B} \\ \midrule
CoT-SC, five samples & 5 & \textbf{76.7} & \textbf{71.1} & \textbf{70.3} & \textbf{67.6} & \textbf{64.4} & \textbf{62.7} & \textbf{66.8} & \textbf{65.1} & \textbf{63.3} \\
Pre-consultation votes & 4 & 69.2 & \underline{67.8} & 66.9 & \underline{65.4} & \underline{62.5} & \underline{62.1} & 64.7 & 62.8 & \underline{61.6} \\
Three independent CoT runs & 3 & \underline{70.7} & 67.2 & 66.7 & -- & -- & -- & 62.8 & 61.6 & 60.1 \\
Two different models & 2 & 70.3 & 63.4 & -- & -- & -- & -- & 61.7 & \underline{63.7} & -- \\
Option-shuffled voting & 3 & 58.1 & 63.9 & 62.8 & -- & -- & -- & 61.3 & 62.5 & 58.2 \\
Three consultation runs & 27 & 67.0 & 65.7 & \underline{70.0} & -- & -- & -- & \underline{65.0} & 61.0 & 60.6 \\
Vote stability & 9 & 66.9 & 64.0 & 65.3 & 63.1 & 59.0 & 60.2 & 61.9 & 58.3 & 59.5 \\
Self-Refine verdict & 2 & 61.5 & 62.3 & 50.5 & -- & -- & -- & 59.5 & 60.5 & 50.1 \\
Post-consultation votes & 9 & 54.3 & 57.3 & 57.2 & 53.6 & 55.0 & 56.1 & 53.3 & 56.1 & 56.7 \\
Three repeated zero-shot runs & 3 & 54.4 & 50.9 & 53.5 & -- & -- & -- & 53.0 & 51.2 & 54.4 \\
MDAgents complexity & 1 & 52.9 & 52.4 & 48.4 & -- & -- & -- & 51.2 & 49.6 & 49.2 \\
 
\dbottomrule\end{tabular}
\begin{tablenotes}[flushleft]\footnotesize
\item[] Calls, model calls per question. Every signal is scored against the final answer of the protocol, so the AUROC of CoT-SC here differs from its AUROC for its own answers in Table~\ref{tab:selective}. The ranking is shown in Fig.~\ref{fig:league}. Two different models, agreement of the zero-shot answers of Qwen3-32B and Llama-3.3-70B. Three consultation runs, agreement of the final answers of three runs of the protocol. Option-shuffled voting and three repeated zero-shot runs, agreement among three zero-shot answers with and without shuffled options. Self-Refine verdict, the self-critique verdict of Self-Refine. MDAgents complexity, the question complexity that MDAgents assigns. Vote stability, no agent changed its vote between steps~2 and~4. On the full VM14K set, raw outputs of Llama-3.1-8B are available only from the topic-aware variant (one seed), from which its vote-based signals are taken. On VM14K, only the four signals available under the unified configuration were computed. --, not available. Bold, best signal in each column. Underlined, second best.
\end{tablenotes}
\end{threeparttable}
\end{table*}

\begin{table*}[pos=tbp]\centering\small
\begin{threeparttable}
\caption{Sensitivity analyses.}\label{tab:sens}
\setlength{\tabcolsep}{4pt}
\begin{tabular}{@{}>{\raggedright\arraybackslash}p{28mm}>{\raggedright\arraybackslash}p{54mm}rrrrrr@{}}\dtoprule
\multicolumn{2}{@{}l}{} & \multicolumn{2}{c}{Unanimous votes (\%)} & \multicolumn{2}{c}{Unanimous and wrong (\%)} & \multicolumn{2}{c}{AUROC} \\
\cmidrule(lr){3-4}\cmidrule(lr){5-6}\cmidrule(l){7-8}
\multicolumn{2}{@{}l}{Analysis} & Before & After & Before & After & Before & After \\ \midrule
\multirow[t]{9}{=}{(a) Run-time votes only} & MedQA-USMLE, Qwen3-32B & 79.6 & 96.5 & 8.0 & 15.0 & 69.3 & 54.3 \\
 & MedQA-USMLE, Llama-3.3-70B & 87.8 & 95.5 & 5.1 & 7.6 & 68.3 & 57.5 \\
 & MedQA-USMLE, Llama-3.1-8B & 46.0 & 81.0 & 9.0 & 23.4 & 64.8 & 57.3 \\
 & VM14K, Qwen3-32B & 61.6 & 94.7 & 12.2 & 25.8 & 63.8 & 54.0 \\
 & VM14K, Llama-3.3-70B & 74.8 & 88.7 & 16.9 & 25.2 & 64.3 & 54.5 \\
 & VM14K, Llama-3.1-8B & 35.8 & 56.2 & 12.5 & 25.9 & 62.0 & 56.0 \\
 & VM14K, heterogeneous council & 38.5 & 48.6 & 5.1 & 12.9 & 68.5 & 54.9 \\
 & VM14K, consultation without report & 62.6 & 91.4 & 12.9 & 23.5 & 62.0 & 55.1 \\
 & VM14K full set, Llama-3.3-70B & 69.6 & 75.8 & 17.7 & 22.9 & 61.4 & 53.9 \\
\midrule
\multirow[t]{3}{=}{(b) Full VM14K set by difficulty, Qwen3-32B\tnote{a}} & Easy (3,556) & 69.7 & 93.3 & 14.3 & 24.9 & 65.8 & 53.7 \\
 & Medium (6,315) & 67.7 & 94.6 & 15.1 & 27.4 & 64.3 & 53.1 \\
 & Challenging or hard (1,223) & 60.7 & 94.4 & 15.0 & 30.7 & 63.0 & 52.9 \\
\midrule
\multirow[t]{4}{=}{(c) Candidate label errors excluded} & MedQA-USMLE, Qwen3-32B & 81.1 & 96.8 & 7.1 & 13.9 & 71.1 & 54.7 \\
 & MedQA-USMLE, Llama-3.3-70B & 90.4 & 96.1 & 4.4 & 6.5 & 70.2 & 58.5 \\
 & VM14K full set, Qwen3-32B\tnote{a} & 67.4 & 94.1 & 13.8 & 25.9 & 65.6 & 53.5 \\
 & VM14K full set, Llama-3.3-70B & 77.5 & 90.2 & 17.0 & 24.4 & 63.7 & 56.4 \\
 
\dbottomrule\end{tabular}
\begin{tablenotes}[flushleft]\footnotesize
\item[] Before and after consultation. Unanimous votes and unanimous and wrong are shares of all questions. AUROC of agreement for detecting an incorrect final answer. (a)~Run-time votes only, without re-extraction (Section~\ref{sec:votes}), with a missing vote counted as dissenting. (b)~Full VM14K set by difficulty label, with the number of questions in brackets. (c)~Questions on which at least 90\% of the 27 answers across models and strategies chose the same option other than the reference answer are excluded, 17 of 1,272 on MedQA-USMLE and 175 of 11,094 on the full VM14K set.
\item[a] Run-time votes, which do not depend on re-extraction.
\end{tablenotes}
\end{threeparttable}
\end{table*}

\bibliographystyle{model1-num-names}
\bibliography{refs}

\end{document}